%% file: main.tex
\documentclass[conference]{IEEEtran}
\IEEEoverridecommandlockouts
\usepackage{amsmath,amssymb,amsfonts}
\usepackage{graphicx}
\usepackage{textcomp}
\usepackage[table]{xcolor}
\usepackage{subfig}  
\usepackage{tabularx}
\usepackage{hyperref}
\usepackage{colortbl}
\usepackage{booktabs}
\usepackage{enumerate,enumitem}

\usepackage{algorithm}
\usepackage{algpseudocode}

\usepackage{biblatex}
\def\BibTeX{{\rm B\kern-.05em{\sc i\kern-.025em b}\kern-.08em
T\kern-.1667em\lower.7ex\hbox{E}\kern-.125emX}}

\makeatletter
\newcommand{\linebreakand}{%
  \end{@IEEEauthorhalign}
  \hfill\mbox{}\par
  \mbox{}\hfill\begin{@IEEEauthorhalign}
}
\makeatother

\title{DeepCollide: Scalable Data-Driven High DoF Configuration Space Modeling using Implicit Neural Representations}

\author{
\IEEEauthorblockN{Gabriel Guo}
\IEEEauthorblockA{\textit{Department of Computer Science} \\
\textit{Columbia University}\\
New York, NY, USA \\
g.guo@columbia.edu}
\and
\IEEEauthorblockN{Judah Goldfeder}
\IEEEauthorblockA{\textit{Department of Computer Science} \\
\textit{Columbia University}\\
New York, NY, USA \\
jag2396@columbia.edu}
\and
\IEEEauthorblockN{Aniv Ray}
\IEEEauthorblockA{\textit{Department of Computer Science} \\
\textit{Columbia University}\\
New York, NY, USA \\
ar4180@columbia.edu}
\linebreakand
\IEEEauthorblockN{Tony Dear}
\IEEEauthorblockA{\textit{Department of Computer Science} \\
\textit{Columbia University}\\
New York, NY, USA \\
tony.dear@columbia.edu}
\and
\IEEEauthorblockN{Hod Lipson}
\IEEEauthorblockA{\textit{Department of Mechanical Engineering} \\
\textit{Columbia University}\\
New York, NY, USA \\
hod.lipson@columbia.edu}
}

\begin{document}

\maketitle

\begin{abstract}
    Collision detection is essential to virtually all robotics applications. However, traditional geometric collision detection methods generally require pre-existing workspace geometry representations; thus, they are unable to infer the collision detection function from sampled data when geometric information is unavailable. Learning-based approaches can overcome this limitation. Following this line of research, we present DeepCollide, an implicit neural representation method for approximating the collision detection function from sampled collision data. As shown by our theoretical analysis and empirical evidence, DeepCollide presents clear benefits over the state-of-the-art, as it relates to time cost scalability with respect to training data and DoF, as well as the ability to accurately express complex workspace geometries. We publicly release our code.
\end{abstract}

\input{intro}
\input{related_works}
\input{methods}

\input{theoretical_comparison}
\input{experimental_setup}
\input{results}
\input{discussion_conclusion}

\printbibliography


\end{document}

%% file: intro.tex
\section{Introduction}

\begin{figure*}
    \centering
    \includegraphics[width=0.9\linewidth]{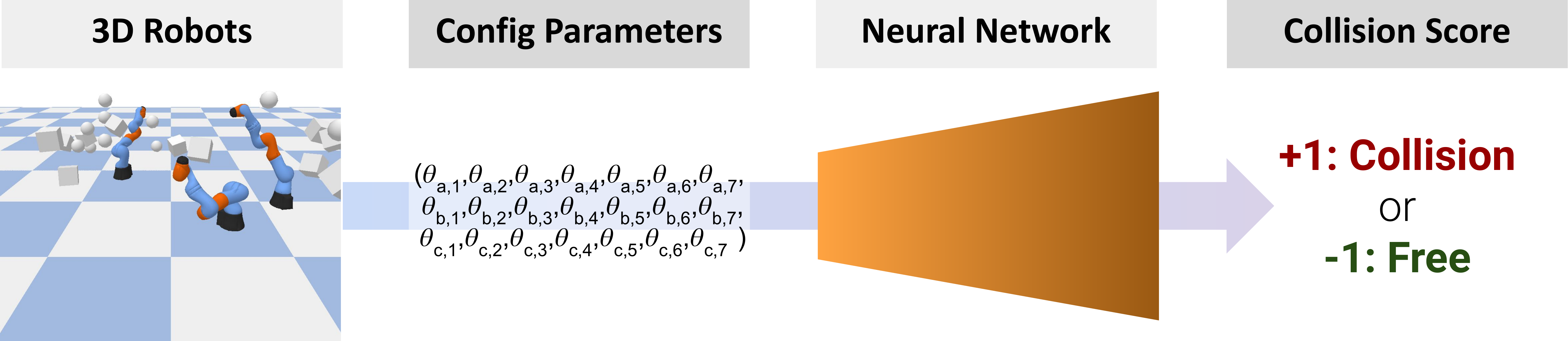}
    \caption{\textbf{System Diagram}: Given robots in a 3-dimensional workspace, DeepCollide can learn the collision detection function based on the joint configurations.}
    \label{fig:schematic}
\end{figure*}

\textbf{Motivation and Background:} Collision detection is crucial to virtually all robotics applications, as it enables robots to find safe paths to their goal states. Typically, collision detection is framed as a binary function: the domain is all possible points in the robot's configuration space (C-space), which comprises of the concatenated degrees of freedom (DoF) \cite{lozano1990spatial, wise2000survey}; the codomain is a binary variable that indicates whether or not the robot collides with an obstacle (including itself) \cite{fastron_das2020learning}.

\textbf{Problem Statement:} 
However, traditional collision detection algorithms (\textit{e.g.}, GJK \cite{gjk_gilbert1988fast}, Lin-Canny \cite{lin_canny_1991fast}, OBBTrees \cite{hierarchial_gottschalk1996obbtree}), while typically fast and accurate, have a major shortcoming: they assume that we already have a model of workspace geometry. For instance, GJK and Lin-Canny only work on convex polytopes \cite{gjk_gilbert1988fast, lin_canny_1991fast}, and a crucial step in OBBTrees is to decompose polygons into collections of triangles \cite{hierarchial_gottschalk1996obbtree}.
This limits their usefulness when we have a non-geometric representation of the workspace obstacles, \textit{e.g.}, point clouds or sampled collision data (such as that collected during PRM \cite{prm_kavraki1996probabilistic} or RRT \cite{kuffner2000rrt}).
In particular, we are interested in building collision detection algorithms for when we have sampled collision data, \textit{i.e.}, list of robot configurations paired with whether or not there was a collision. 

This is a worthwhile problem to study, considering the following scenario: 
A business owner has some robots in a warehouse. The geometries and positions of the robots are already known, as the business owner placed them there. However, the layout of the robots' surroundings changes throughout the day, as new packages arrive and get shipped out. There are no cameras in the warehouse, nor are there measurements for every single object that passes through the warehouse. 
In lieu of object geometry information, the primary way to build a collision detection function would be to learn from samples, \textit{i.e.}, learning-based approach.

While there are likely practical limitations to learning a collision detection function from samples in real-life (\textit{e.g.}, knocking objects out of place, having the time to collect all the samples), it is a problem that traditional geometric collision checkers do not readily hold the solution to. Thus, we find it valuable to study this problem, at least in the academic setting.


\textbf{Previous Solutions:} Indeed, many robotics researchers have tried learning-based approaches (\textit{e.g.}, SVM, KNN) \cite{fastron_das2020learning, forward_kinematics_kernel_das2020forward, fastron_verghese2022configuration, pan2016fast, cspace_svm_pan2015efficient, huh2016learning, huh2017adaptive, chase2020neural}. The current state-of-the-art (as it relates to speed, time, and reproducibility) is the open-sourced Fastron \cite{fastron_das2020learning} and its variants \cite{forward_kinematics_kernel_das2020forward, fastron_verghese2022configuration}. 

It is important to note that previous works in learning-based collision detection primarily studied it with the goal of improving inference time over that of geometric collision checkers, and in the test environments presented in their papers, they did indeed achieve this goal \cite{fastron_das2020learning, fastron_verghese2022configuration, forward_kinematics_kernel_das2020forward, cspace_svm_pan2015efficient, pan2016fast, chase2020neural}. However, our own preliminary analyses suggest that the improved inference time was likely due to factors including, but not limited to: (1) primarily testing on low-data and low-DoF settings, when in fact, the methods (SVM \cite{cspace_svm_pan2015efficient}, KNN \cite{knn_burns2005toward}, Fastron \cite{fastron_das2020learning}) have been analytically shown to scale poorly with respect to training dataset size and DoF \cite{scikit-learn}; (2) naive calculation of baseline collision detection time, \textit{e.g.}, counting PyBullet \cite{coumans2016pybullet} overhead time when measuring GJK \cite{gjk_gilbert1988fast, chase2020neural}. 

In contrast, we do not even compare our method to geometric collision detectors, because we aim to solve a different (albeit closely related) problem: learning the collision detection function from sampled data, as opposed to detecting collisions from geometric models. We still compare our method to other learning-based methods.

Based on our experiments, we found two major areas for improvement in learning-based collision detection approaches: (1) the expressivity in modeling complex collision functions is limited, due to the lack of learned feature transformations (\textit{e.g.}, Fastron uses a Gaussian kernel \cite{fastron_das2020learning}, which was found to be sub-optimal in a follow-up work \cite{forward_kinematics_kernel_das2020forward}); (2) the time cost of previous approaches (particularly Fastron) does not scale well with training data or DoF.

\textbf{Design Goals:} To overcome these areas for improvement, we have two design goals:
\begin{enumerate}
    \item \textbf{Expressiveness:} The proposed model should be able to approximate the collision detection function in a wide variety of complex workspace geometries.
    \item \textbf{Scalability:} The proposed model's time complexity should scale favorably (linear or quicker) as a function of dataset size and DoF.
\end{enumerate}

\textbf{Our Approach:} Here, we present DeepCollide, an expressive, scalable neural network method for approximating the collision detection function. DeepCollide builds on the recent success of implicit neural representations \cite{park2019deepsdf, tancik2020fourier, sitzmann2019scene, sitzmann2020implicit, mildenhall2021nerf} -- which are neural networks that parameterize signals (\textit{e.g.}, SDF, volumetric rendering, colors in an image) as continuous functions \cite{sitzmann_awesome_implicit_neural_representations}. DeepCollide uses many state-of-the-art techniques, such as the forward kinematics kernel \cite{forward_kinematics_kernel_das2020forward}, sinusoidal positional encoding \cite{mildenhall2021nerf, tancik2020fourier}, and skip connections \cite{park2019deepsdf, resnet_he2016deep}.

\textbf{Results:} We show through Big-O runtime analysis as a function of training dataset size that DeepCollide achieves constant time complexity per inference and linear time complexity over the training cycle, which outperforms Fastron FK (linear inference time, near-quadratic training time) \cite{fastron_das2020learning, forward_kinematics_kernel_das2020forward}, the current state-of-the-art. Then, we empirically show in simulation that DeepCollide achieves state-of-the-art results with respect to the runtime vs. error tradeoff. Finally, we empirically investigate the impact of DoF (up to $42$), collision density, and training dataset size (up to $100,000$ training points) on the performance of learning-based collision detection methods, and find that ours performs favorably, no matter which factors are varied.

\textbf{Summary of Contributions:}

\begin{itemize}
    \item We design DeepCollide, a scalable, lightweight neural network that approximates the collision detection function in high-DoF (up to $42$) settings.
    \item We conduct a theoretical analysis to show that DeepCollide outperforms the state-of-the-art learning-based collision detection method in regards to scalability of speed with training data. 
    \item We provide empirical evidence that DeepCollide again outperforms the state-of-the-art in terms of both speed and correctness, across a variety of high-DoF settings.
    \item We investigate the impact of various factors -- DoF, collision density, sample size -- on the performance of learning-based collision detection methods.
    \item To promote reproducibility, we release our code at \url{https://github.com/gabeguo/RobotConfigSpaceGen/tree/revision/_GenerateEnvironment}
\end{itemize}

%% file: related_works.tex
\section{Related Works}

\subsection{Geometric Collision Detection Methods}

Typically, the Gilbert–Johnson–Keerthi algorithm (GJK) is used for quick collision detection \cite{gjk_gilbert1988fast, ong2001fast}. One requirement of the algorithm is that all colliding objects must be convex -- to deal with this, practitioners can either use a convex approximation to the objects (\textit{e.g.}, bounding box, bounding cylinder) \cite{cesati1995parameterized, bounding_box_barequet2001efficiently, bounding_box_huebner2008minimum, wise2000survey, coumans2016pybullet}, or decompose the objects as collections of convex objects \cite{hierarchical_lozano1979algorithm}. Additionally, while it is not technically a collision detection method, the expanding polytopes algorithm (EPA) is frequently used in conjunction with GJK to calculate penetration depth when collisions do happen \cite{epa_van2001proximity, coumans2016pybullet}.
Another geometric collision detection algorithm is the Lin-Canny algorithm, which also requires that objects be convex \cite{lin_canny_1991fast}. Hierarchical data structures, such as OBBTrees \cite{hierarchial_gottschalk1996obbtree}, have also been proposed for use in collision checking, and have been implemented in the Flexible Collision Library (FCL) \cite{pan2012fcl}. 

A disadvantage of geometric collision checking methods is that they require a pre-existing representation of obstacle and workspace geometry, which may prove problematic if we only have sampled collision data, \textit{i.e.}, list of configurations with their associated collision statuses. In contrast, learning-based methods can work given only sampled collision data, and they can also work on pre-existing obstacle geometry representations (by using the representations to generate training samples).
Another disadvantage of geometric methods (particularly GJK) is that their time costs increase with the number of obstacles \cite{fastron_das2020learning, forward_kinematics_kernel_das2020forward}.
In this work, our focus is on learning-based methods, so we do not benchmark against geometric methods.

\subsection{Learning-Based Collision Detection}

Prior works in learning-based collision detection use a variety of machine learning models (SVM, GMM, KNN) and sampling methodologies
\cite{cspace_svm_pan2015efficient, knn_burns2005toward, han2020configuration, pan2016fast, huh2016learning, huh2017adaptive}. 
However, one of their main limitations is that the highest DoF explored was $12$ \cite{knn_burns2005toward}.
Thus, one of the major contributions of our work is that we extend the data-driven method to work in DoF $\geq 14$. 

Other works explore neural networks in collision detection \cite{garcia2002neural, chase2020neural}. However, these works do not test on a wide variety of environments, nor do they open-source their code for fair comparison. In contrast, our evaluation is much more extensive (investigating impact of DoF, collision density, and training set size; on many randomly generated environments), and we open-source our code.

Recently, Fastron \cite{fastron_das2020learning} and its variants \cite{fastron_verghese2022configuration, forward_kinematics_kernel_das2020forward} have shown state-of-the-art performance in learning-based collision detection. The authors \cite{fastron_das2020learning} compared their method against SSVM \cite{ssvm_huang2010sparse}, ISVM \cite{cspace_svm_pan2015efficient}, FCL \cite{pan2012fcl}, and GJK \cite{gjk_gilbert1988fast} -- and showed that their method was superior to its predecessors, as it relates to the overall speed-error tradeoff (typically, other methods were orders of magnitude slower, with only a marginal reduction in error). Later on, Fastron with the forward kinematics kernel (Fastron FK) was shown to outperform Fastron, in both speed and error \cite{forward_kinematics_kernel_das2020forward}. 
However, even as Fastron FK and Fastron improve over other methods, they still only evaluate on a maximum of 7 DoF, with only a few thousand training points \cite{fastron_das2020learning, forward_kinematics_kernel_das2020forward}.

Another major limitation of existing machine learning methods is that their training and/or inference time greatly increases with dataset size and number of features (\textit{i.e.}, DoF): for instance, SVM \cite{svm_original_cortes1995support} is known to have $O(n_{features} \times n_{samples}^{3})$ training time complexity, while KNN (KDTree implementation) \cite{knn_original_cover1967nearest, kdtree_original_bentley1975multidimensional} is known to have $O(n_{features} \times \text{log }n_{samples})$ time cost per inference \cite{scikit-learn}. We also show in Section \ref{sec:theoretical} that Fastron does not scale favorably either, with almost $O(n_{features} \times n_{samples})$ time cost per inference and $O(n_{features}^{2} \times n_{samples})$ training time complexity (although some optimizations are made, which result in slightly faster performance in practice) \cite{fastron_das2020learning}.
In contrast, our deep learning method is able to achieve $O(n_{samples} \times n_{features})$ training time complexity, and $O(n_{features})$ inference time complexity (empirically, we find that $n_{features}$ provides a negligible runtime contribution, such that the runtime almost entirely depends on $n_{samples}$).

Nevertheless, seeing as Fastron FK outperformed Fastron, which outperformed previous state-of-the-art learning-based methods, we only benchmark our method against Fastron FK. 
Furthermore, the theoretical runtime analysis above and in Section \ref{sec:theoretical} suggests that our method would outperform the others anyways.
Another reason for only bechmarking against Fastron FK is that based on our inspection, it seems that Fastron is the only major open-sourced learning-based method \cite{fastron_das2020learning} (and calculating the forward kinematics kernel is easy with PyBullet \cite{coumans2016pybullet}) \footnote{While SVM, KNN, and GMM are open-sourced via SciKit-Learn \cite{scikit-learn}; the works that use these machine learning algorithms \cite{knn_burns2005toward, cspace_svm_pan2015efficient, huh2016learning, huh2017adaptive} alter the methods in meaningful ways, such that using the SciKit-Learn implementations is probably not a fair comparison.}, so we can be confident that we are conducting a fair comparison.

\subsection{Implicit Neural Representations}

Implicit neural representations are neural networks that parameterize signals as continuous functions \cite{sitzmann_awesome_implicit_neural_representations}. Commonly, they take in $(x, y, z)$ coordinates as input, and output the function value at that point; but, any arbitrary coordinate system (\textit{e.g.}, polar, spherical, 6D) can be used as input. They have shown recent state-of-the-art results in a variety of applications relevant to the robot learning community: signed distance functions \cite{park2019deepsdf}, volumetric rendering \cite{mescheder2019occupancy, sitzmann2019scene, mildenhall2021nerf, tancik2020fourier, sitzmann2020implicit}, and robot self-modeling \cite{selfmodeling_chen2022fully}. We build on the success of these works by extending the implicit neural representation approach to modeling the collision detection function.

%% file: methods.tex
\section{Methods}

\begin{figure}
    \centering
    \includegraphics[width=0.75\linewidth]{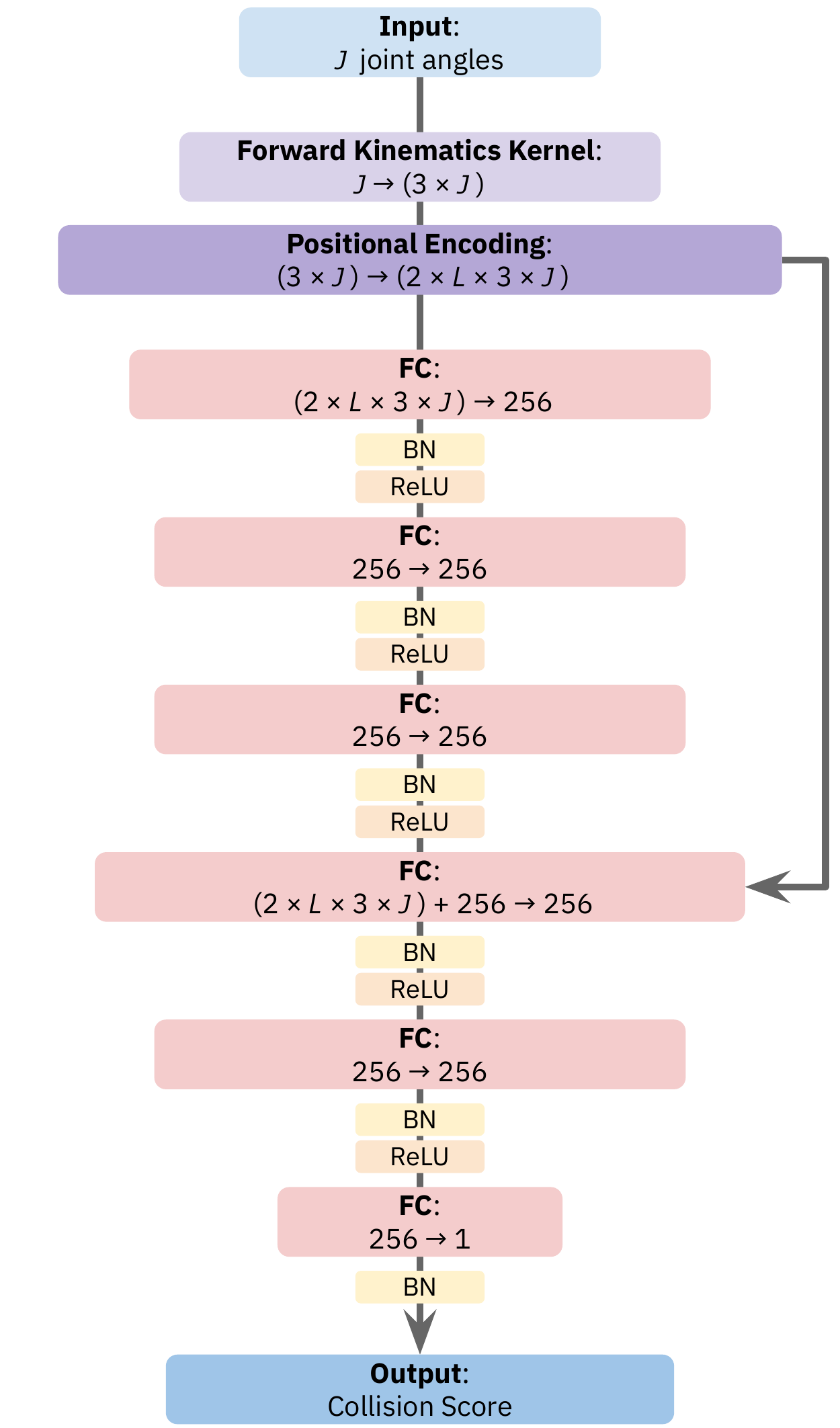}
    \caption{\textbf{DeepCollide Architecture}: We leverage multiple state-of-the-art techniques, including: forward kinematics kernel \cite{forward_kinematics_kernel_das2020forward}, positional encoding \cite{mildenhall2021nerf}, BatchNorm \cite{ioffe2015batchnorm}, and skip connections \cite{resnet_he2016deep}. $J$ is the number of joint angles, and $L$ is the number of frequencies. }
    \label{fig:architecture}
\end{figure}

\subsection{Design Goals}

We design DeepCollide to have the following properties:
\begin{enumerate}
    \item \textbf{Expressiveness:} It should be able to approximate the collision detection function in a wide variety of complex scene geometries, ranging from low collision density to high collision density. 
    \item \textbf{Scalability:} Its speed and correctness should persist even in high-DoF scenarios. Furthermore, when presented with more training data, DeepCollide should be able to increase its correctness \textit{without sacrificing inference speed}, and training speed should only scale \textit{linearly} with the amount of training data presented. 
\end{enumerate}

\subsection{DeepCollide Architecture}

The input to DeepCollide is $J$ joint angles, each representing one of the DoF of the robot arm \footnote{In our paper, we evaluate robot arms, following the convention in \cite{fastron_das2020learning, forward_kinematics_kernel_das2020forward}, but the method can easily be extended to other robot geometries.}. The output of DeepCollide is a collision score, where a negative number indicates free space ($-1$), and a positive number indicates collision ($+1$), following the convention from Fastron \cite{fastron_das2020learning}. See Figure \ref{fig:architecture}.

\noindent\textbf{Forward Kinematics Kernel:}
The first layer of our model is the forward kinematics kernel, as introduced by Das \textit{et al.} \cite{forward_kinematics_kernel_das2020forward}. The forward kinematics kernel transforms the raw joint angles into control point coordinates (which we choose to be the centers of mass of the robot limbs) in world space (note that this requires knowledge of the robot geometry a priori). Das \textit{et al.} had previously shown the superiority of this representation over the raw joint angles \cite{forward_kinematics_kernel_das2020forward}, and during our model design phase, we came to the same conclusion. The reason for this is that the forward kinematics kernel is a semantically meaningful representation of the configuration, in which two similar configurations | as it relates to collision status | will also have similar representation values, \textit{i.e.}, close $(x, y, z)$ coordinates. 

\noindent\textbf{Positional Encoding:}
We also use a positional encoding (which can be thought of as an approximate representation of the Fourier series frequencies of the collision signal), following previous work in implicit neural representations \cite{mildenhall2021nerf, tancik2020fourier}. Mathematically, 

\begin{equation}
\begin{split}
    e(p) = (sin(\sigma p), cos(\sigma p), sin (2\sigma p), cos (2\sigma p), \ldots, \\ sin(L\sigma p), cos(L\sigma p)) 
\end{split}
\end{equation}
where $p$ is the inputted position (from the forward kinematics kernel), $L$ is the number of frequencies, and $\sigma$ is the frequency step width. This works because it allows our model to learn features of different frequencies, akin to a Fourier series.

We are aware that the state-of-the-art positional encoding technique, proposed by Tancik \textit{et al.}, uses random Fourier features with off-axis frequencies (achieved by taking a linear combination of the inputted coordinates) \cite{tancik2020fourier}, while ours only uses on-axis frequencies. Empirically, we found our method to work better for the stated problem of collision detection. We believe this is so because while Tancik \textit{et al.} only sought to parameterize signals with two or three coordinates, our signals have up to $42 \times 3 = 126$ coordinates (in the $42$ DoF case) | this makes it computationally prohibitive to get a dense enough random sampling of the frequencies that make up the Fourier series of the collision boundary signal. 

\noindent\textbf{Skip Connection:}
Following the state-of-the-art \cite{park2019deepsdf, mildenhall2021nerf, densenet_huang2017densely}, we use a skip connection from the positional encoding layer to a fully-connected layer near the end of the network. This allows information flow throughout the model. We make the skip connection concatenative, rather than additive, so that both the low-level and high-level representations of the information are preserved. 

\noindent\textbf{BatchNorm:}
After every fully-connected layer, we use BatchNorm \cite{ioffe2015batchnorm}. Empirically, we find that this makes it much easier to train a deep model, without having to worry about exploding or vanishing gradients. We also find that it is helpful to add BatchNorm at the end of the model (right before the output). We believe this is because the decision boundary for whether or not there is a collision lies at $\mathcal{F}(x) = 0$, and by default, BatchNorm's tendency is to normalize the output to be centered at $0$.

\noindent\textbf{Other Architectural Choices:}
We have five fully-connected hidden layers in our network, as networks with many hidden layers have consistently been shown to perform better across a variety of tasks \cite{resnet_he2016deep, mildenhall2021nerf, park2019deepsdf, selfmodeling_chen2022fully}. However, there is also a size vs. speed tradeoff, and since speed is paramount in collision detection, we do not further increase the size.

We also make our output a scalar (\textit{i.e.}, one-dimensional) collision score rather than a binary output (\textit{i.e.}, classification of colliding or not colliding), as we empirically found it to be easier to train than a binary output. We think this is because regressing a scalar allows the model to learn a smoother transition from collision to free space, rather than choosing between disjoint classes.


\subsection{DeepCollide Training}

\noindent\textbf{Training Loop Details:}
Our target outputs are $-1$ for free space and $+1$ for collision, following the convention in \cite{fastron_das2020learning}. We use L1 loss. We train our model to a maximum of $50$ epochs, subject to early stopping. We use the Adam optimizer with $\beta = (0.9, 0.999)$ \cite{kingma2014adam}; and a learning rate of $10^{-3}$, which decays to $10^{-7}$ on the $50^\text{th}$ epoch following a cosine decay schedule \cite{cosine_loshchilov2016sgdr}. We use $95\%$ of the training data to optimize the model weights, and $5\%$ of the training data to validate what the best epoch to take our model from is \footnote{In the deep learning literature, these are referred to as the training and validation sets, and are separate from the testing set. Here, we collectively refer to them both as the training data, because prior works in configuration space approximation do not use a validation set \cite{fastron_das2020learning, forward_kinematics_kernel_das2020forward}}.

\noindent\textbf{Bias Scaling:}
We also have a bias parameter, $\beta$, which controls the weighting we give to positive samples. Essentially, as noted in the state-of-the-art collision detection works, it is often desirable to build a model that is better at detecting more collisions, at the expense of missing some of the free space \cite{fastron_das2020learning, forward_kinematics_kernel_das2020forward}.

To this end, we multiply all the training labels with value $+1$ (\textit{i.e.}, collisions) by $\beta$, while leaving the labels with value $-1$ (\textit{i.e.}, free space) unchanged. Since we are using a regression-based approach, this increases the gradient of the objective function for collisions when $\beta > 1$, thereby encouraging the model to disproportionately learn how to predict collisions. When $\beta = 1$, the model is biased neither towards collisions or free space, and when $\beta < 1$, the model is biased towards free space.

%% file: theoretical_comparison.tex
\section{Theoretical Analysis}\label{sec:theoretical}

Here, we show how DeepCollide scales more favorably in theoretical time complexity than Fastron \cite{fastron_das2020learning}, when considering training dataset size and DoF together. (Analysis on methods like SVM and KNN can be found in the SciKit-Learn online manual \cite{scikit-learn}, which shows that they do not scale as favorably as we would like.)

\subsection{Training Time}\label{subsec:train_theoretical_analysis}

Firstly, we analyze how the total training time scales with respect to training dataset size and DoF.

\noindent\textbf{DeepCollide:}
Algorithm \ref{alg:nn_training_loop} shows a simplified version of our neural network training loop. The outer loop itself has $O(E)$ time complexity (one iteration per epoch), and the inner loop itself has $O(\frac{|T|}{B})$ time complexity (one iteration per batch). In the sequential computation case, the statements in the inner loop have $O(B * |\mathcal{F}|)$ time complexity (lines \ref{alg:forward_pass}, \ref{alg:backprop}, and \ref{alg:gradient_descent} are $O(B * |\mathcal{F}|)$; line \ref{alg:loss} is $O(B)$), where $|\mathcal{F}|$ is the number of parameters to the network, as they batch-operate on $B$ items at a time from the dataset by passing them through the neural network. Thus, combined, we have overall runtime complexity in the sequential computation case of $O(E) * O(\frac{|T|}{B}) * O(B * |\mathcal{F}|) = O(E * |T| * |\mathcal{F}|)$. 

We note that with GPU-enabled parallelism, the statements in the inner loop only have $O(|\mathcal{F}|)$ runtime complexity, which gives us $O(\frac{E * |T| * |\mathcal{F}|}{B})$ overall runtime complexity. In our work, we indeed use GPUs to train (but not to test).

Going deeper into the scalability of $|\mathcal{F}|$, we see from Figure \ref{fig:architecture} that $|\mathcal{F}|$ is actually $O(d * L)$, where $d$ is the number of DoF and $L$ is the number of Fourier series frequencies -- this is due to the positional encoding, which connects to the first and fourth (via a skip connection) fully connected layers. 
So, this results in a sequential runtime complexity of $O(E * |T| * d * L)$, and a parallelized runtime complexity of $O(\frac{E * |T| * d * L}{B})$.

Furthermore, we can consider the number of epochs, batch size, and number of Fourier series frequencies to be constants; since the purpose of this analysis is to understand how runtime scales with respect to training dataset size and DoF. Under these assumptions, the runtime complexity ends up as $O(|T| * d)$
\footnote{We find that in practice, the number of DoF $d$ plays a negligible role in DeepCollide's runtime complexity (see Section \ref{sec:dof_impact_empirical}).}.

\begin{algorithm}
\caption{Neural Network Training Loop.}
\begin{algorithmic}[1]
\Procedure{TrainLoop}{Neural Network $\mathcal{F}$, Training Data $T$, Epoch Count $E$, Batch Size $B$, Learning Rate $R$, Optimizer $O$, Loss Function $L$} 
  \For{$i = 1 \ldots E$}
    \For{$j \in [0, B, 2B, \ldots, \text{floor(}\frac{|T|}{B}\text{)}*B]$}
        \State $t = T[j:j+B]$
        \State $y = \mathcal{F}.\text{forward}(t.\text{configurations})$ \label{alg:forward_pass}
        \State $l = L(y, t.\text{labels})$ \label{alg:loss}
        \State $g = \mathcal{F}$.backprop($l$) \label{alg:backprop}
        \State $O$.gradientDescent($\mathcal{F}$, $g$, $R$) \label{alg:gradient_descent}
    \EndFor
  \EndFor
\EndProcedure
\end{algorithmic}
\label{alg:nn_training_loop}
\end{algorithm}

\noindent\textbf{Fastron:}
Fastron's prediction function is $sign(f(\mathbf{x}))$, where positive values indicate collision, and negative values indicate free space. More specifically, 
\begin{equation}\label{eqn:fastron_inference}
f(\mathbf{x}) = \Sigma_{i} \langle \phi(\mathcal{X}_i), \phi(\mathbf{x}) \rangle _{\mathcal{F}}\alpha_{i},
\end{equation}
where $\mathbf{x}$ is the query configuration, $i$ is the index of a training configuration, $\mathcal{X}_{i}$ is a training configuration, $\alpha_{i}$ is the weight given to that training configuration, and $\phi: \mathcal{R}^{d} \rightarrow \mathcal{F}$ maps the $d$-dimensional configuration to another feature space with a rational quadratic kernel (see \cite{fastron_das2020learning} for more details). The idea is that we measure the similarity between each query point and the already seen points, and weight each comparison according to its importance and collision status.

The learnable parameters in Fastron are the $\alpha_{i}$ (weights for each training sample). Thus, in training,
Fastron updates $\alpha$ to satisfy the condition: $\mathbf{y}_{i} * (\mathbf{K \alpha})_{i} > 0$, where $\mathbf{y}_{i}$ is the ground truth label for training sample $i$, and $\mathbf{K}$ is the Gram matrix (can be thought of as similarity scores) with $K_{ij} = \langle \phi(\mathcal{X}_i), \phi(\mathcal{X}_j) \rangle$ \cite{fastron_das2020learning}.
Thus, if calculated naively, the training time complexity of Fastron would be $O(|T|^{2} d)$, where $|T|$ is the size of the dataset and $d$ is the number of DoF, since we would need to calculate every entry of the $|T| \times |T|$ Gram matrix; which requires taking an inner product on $d$-dimensional data (with the kernel trick for a rational quadratic kernel function \cite{fastron_das2020learning, williams2006gaussian}). 

However, Fastron uses lazy Gram matrix evaluation, in which at most one column (\textit{i.e.}, the column whose corresponding training point was found to have the most negative margin in $\mathbf{y} \odot (\mathbf{K \alpha})$) is calculated per optimization iteration. Furthermore, the authors impose an iteration cap $\mathcal{I}_{max}$. In practice, this means that the runtime is $O(\mathcal{I}_{max} |T| d)$, since we have $O(\mathcal{I}_{max})$ iterations, in which one Gram matrix column with $O(|T|)$ entries is computed, where each entry has $O(d)$ cost to calculate. We do not treat $\mathcal{I}_{max}$ as a constant, because to optimize the model in practice, $\mathcal{I}_{max}$ should approach $|T|$.

Further optimizations are made, such as the support point cap (which we did not analyze, for simplicity and length), but they do not significantly affect the overall conclusion of the runtime analysis. More details can be found in \cite{fastron_das2020learning}.

\subsection{Inference Time}

Now, we wish to analyze how the time per inference scales with respect to training dataset size and DoF. We analyze this rate rather than the total inference time over the whole testing set, because the idea for learning-based methods is that we can make an arbitrary number of collision status inferences.

\noindent\textbf{DeepCollide:}
In inference, DeepCollide only requires a singular forward pass per item. Thus, a singular inference is done in $O(|\mathcal{F}|)$ time. Recall from Section \ref{subsec:train_theoretical_analysis} that $O(|\mathcal{F}|) = O(d * L)$. Considering number of Fourier frequencies as a constant, inference time reduces to $O(d)$.

\noindent\textbf{Fastron:}
Looking at Equation \ref{eqn:fastron_inference}, each inference takes $O(|T| d)$ time, since it is a weighted average of the kernel functions ($O(d)$ time per data point) over all $O(|T|)$ training data points. If we implement the support point cap proposed by the authors \cite{fastron_das2020learning}, runtime is then $O(\mathcal{S}_{max} d)$. We should keep in mind, however, that in practice (and theory), the closer $\mathcal{S}_{max}$ is to $|T|$, the more accurate predictions will be, since we utilize more training data.

%% file: experimental_setup.tex
\section{Experimental Setup}

\input{experimental_setup/environments}
\input{experimental_setup/models_compared}
\input{experimental_setup/metrics}

\subsection{Implementation Details}
\noindent\textbf{Libraries:}
We conduct all experiments in Python 3.9.16. 
For robot simulation and ground truth collision detection, we use PyBullet 3.2.5 \cite{coumans2016pybullet}.
For general computation, we use Numpy 1.24.3 \cite{numpy_harris2020array}.
For our deep learning models, we use PyTorch 1.10.2 \cite{paszke2019pytorch}. For Fastron, we use the official implementation created by the authors \cite{fastron_das2020learning}. 

\noindent\textbf{Hardware:} 
We run almost all experiments on an Intel(R) Xeon(R) Gold 5115 CPU @ 2.40GHz. 
The only exception is that when we train DeepCollide, we use a NVIDIA GeForce RTX 2080 Ti GPU, because GPUs are widely known to accelerate training of deep learning models. We do not use a GPU to test Fastron, because the official implementation does not support that.
However, we still test DeepCollide on the CPU, just as we test Fastron on the CPU. 
We do this because while it may be feasible to train a collision detection model offline with parallelizable GPU resources, when it is deployed, we may be constrained to only using the CPU resources in the robot.

%% file: experimental_setup/environments.tex
\subsection{Environments}

In designing the experimental environments, our goals are:
\begin{enumerate}
    \item To quantify how DeepCollide performs in the speed vs. error trade-off, as compared to the state-of-the-art approach.
    \item To validate DeepCollide's ability to express a wide variety of scene geometries.
    \item To investigate the scalability of DeepCollide, as it relates to DoF.
    \item To investigate the scalability of DeepCollide, as it relates to training dataset size.
\end{enumerate}

In all environments, we use the Kuka LBR iiwa 7 R800 robot arm, which has seven DoF \cite{kuka_lbr_iiwa_r800} \footnote{The last DoF (the end-effector) actually does not affect collision status, since it just rotates a half-sphere. However, we still include it in our model input, because we think it is important to demonstrate that the model can learn when some DoFs are inconsequential.}. 

Towards the first goal of quantifying how DeepCollide performs in the speed vs. error trade-off, we place three Kuka robots ($21$ DoF) in an environment with $25$ obstacles ($13$ cubes, $12$ spheres). We set the Kuka robots at random positions, and we set the obstacles at random positions and orientations. We create three such environments using different random seeds.

Towards the second goal of validating DeepCollide's ability to express a wide variety of scene geometries, we generate environments containing $10, 20, 30, 40, 50, \text{and } 60$ randomly placed obstacles. All of these environments have three robots, which are randomly placed. For each number of obstacles, we test arrangements where the robots are far away from each other (to minimize robot-robot collisions) and close to each other (such that robot-robot collisions are frequent). We use three random seeds, making for a total of $6 \times 2 \times 3 = 36$ environments in this experiment.

Towards the third goal of investigating DeepCollide's scalability with DoF, we generate environments containing between one ($7$ DoF) and four ($28$ DoF) robots, and $25$ obstacles. Again, both the obstacles and robots are randomly placed, and we use three random seeds, making for $4 \times 3 = 12$ total environments in this experiment.

Towards the fourth goal of investigating DeepCollide's scalability with training data, we generate environments as we did in the first experiment, the difference being that we sample more points from these environments. 

\input{experimental_setup/environment_diagrams.tex}

%% file: experimental_setup/environment_diagrams.tex
\begin{figure}
  \subfloat[$71.3\%$ of the c-space is occupied.
  ]
  {\includegraphics[width=0.16\textwidth]{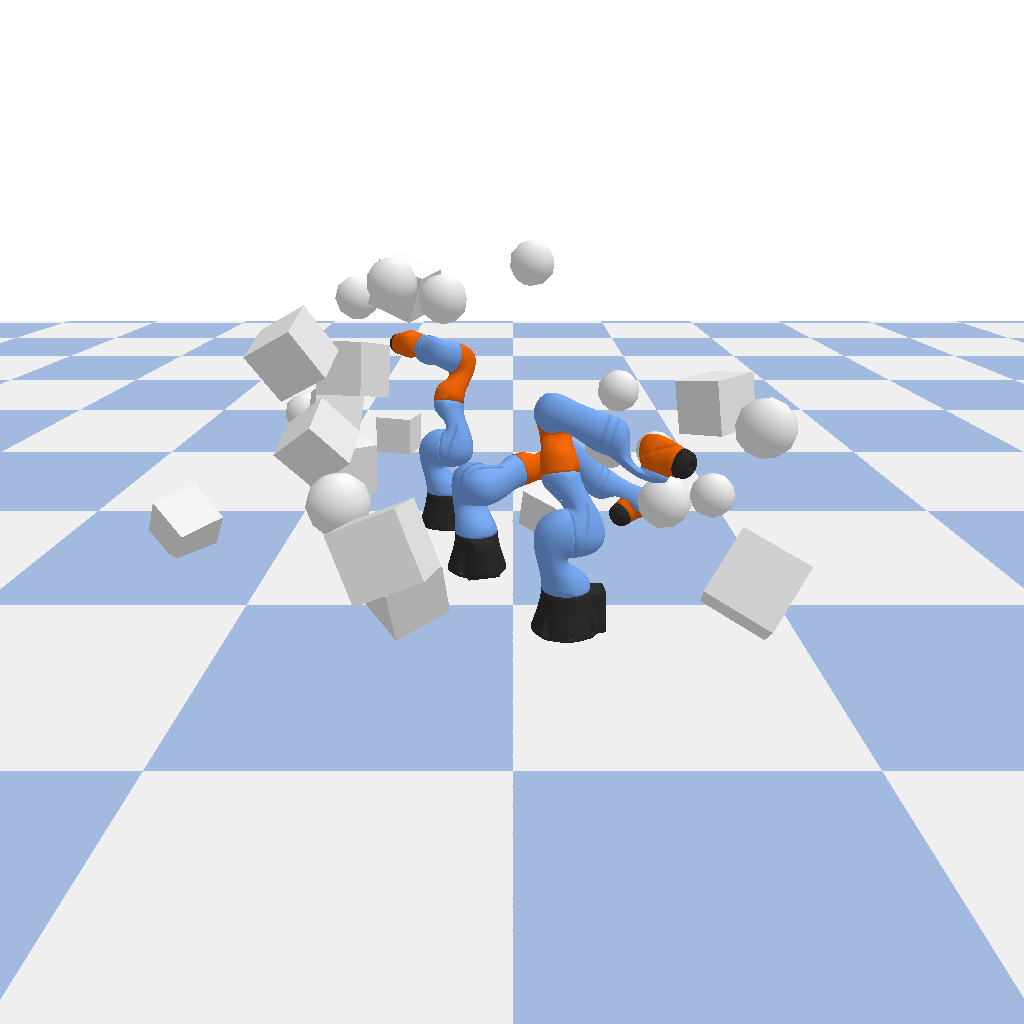}\label{fig:pareto_env0}}
  \hfill
  \subfloat[$47.6\%$ of the c-space is occupied.
  ]
  {\includegraphics[width=0.16\textwidth]{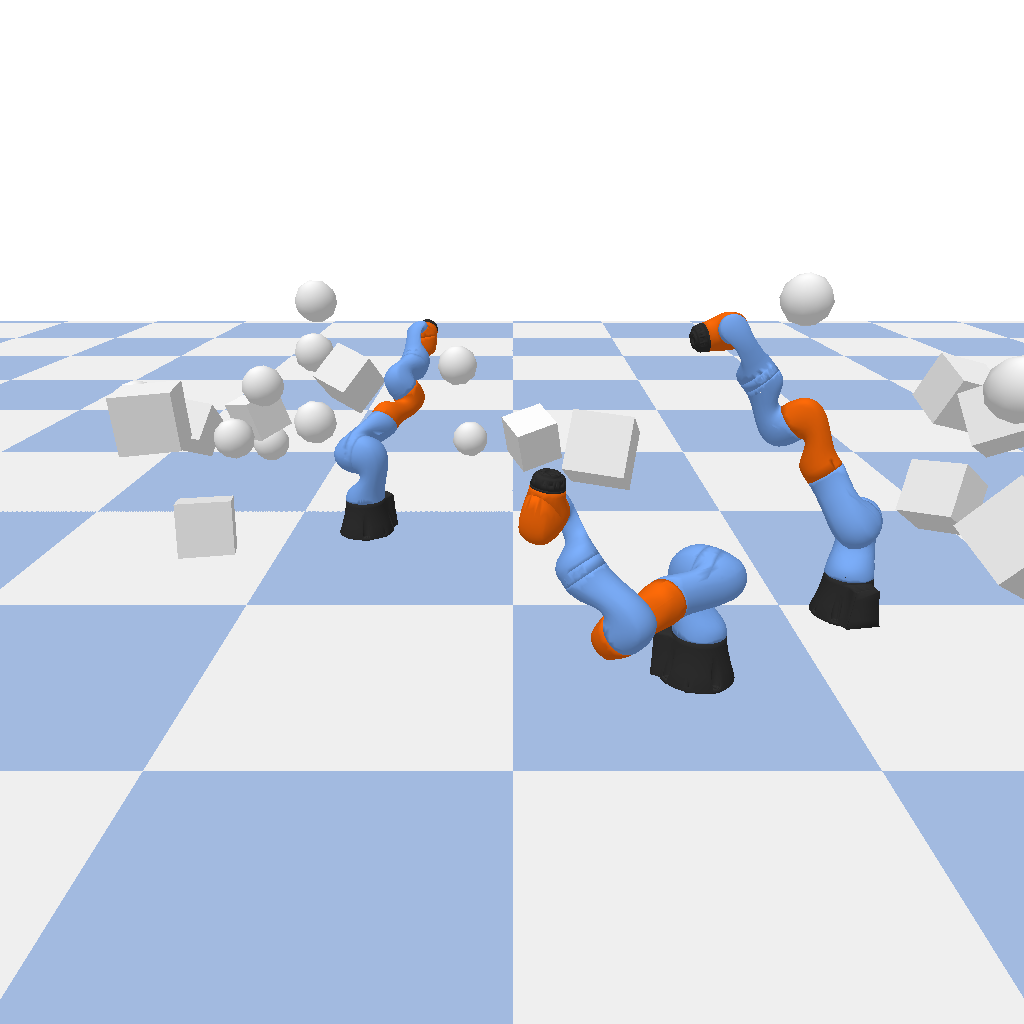}\label{fig:pareto_env1}}
  \hfill
  \subfloat[$66.3\%$ of the c-space is occupied.
  ]
  {\includegraphics[width=0.16\textwidth]{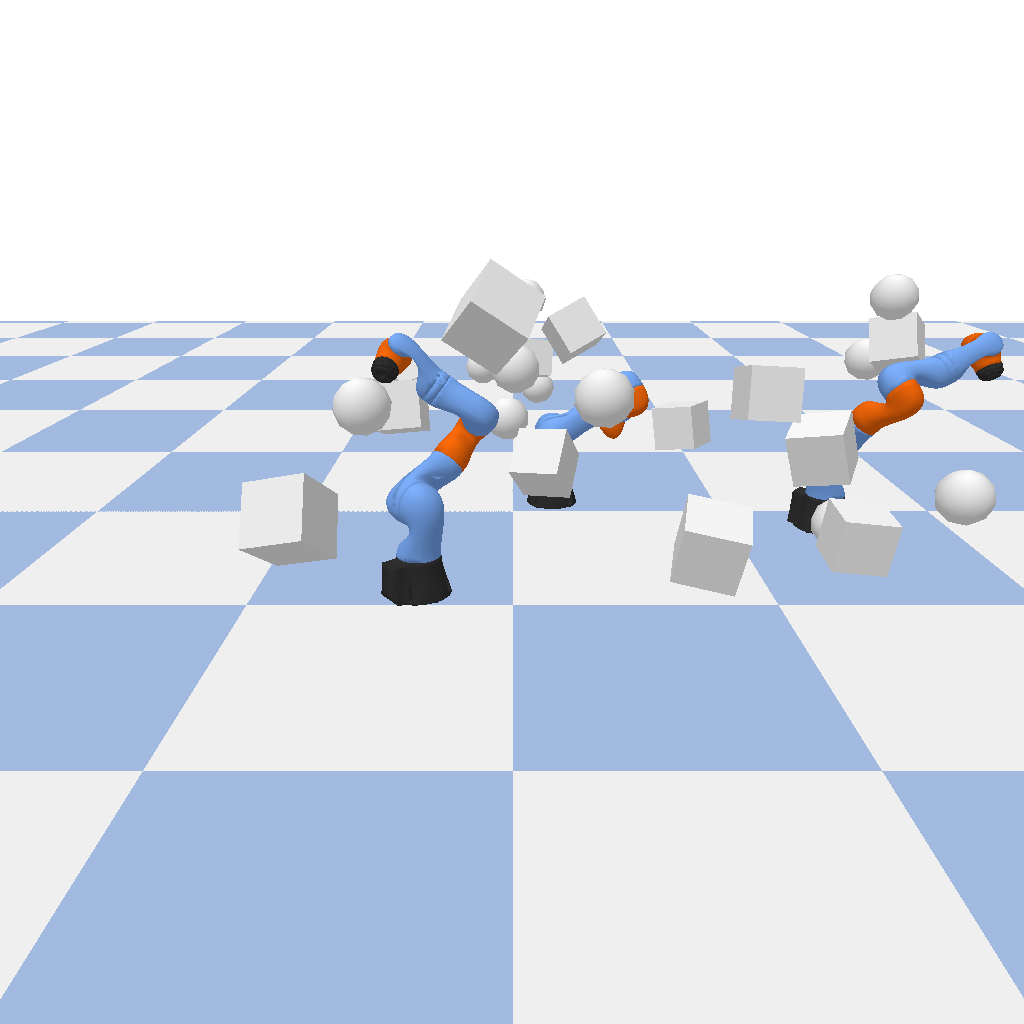}\label{fig:pareto_env2}}
  \caption{\textbf{Environments for Speed vs. Error Analysis.} Each environment has three randomly placed Kuka LBR iiwa 7 R800 robots (21 DoF), and 25 randomly placed obstacles.}
  \label{fig:pareto_envs}
\end{figure}

\begin{figure}
    \subfloat[One robot.]
    {\includegraphics[width=0.16\textwidth]{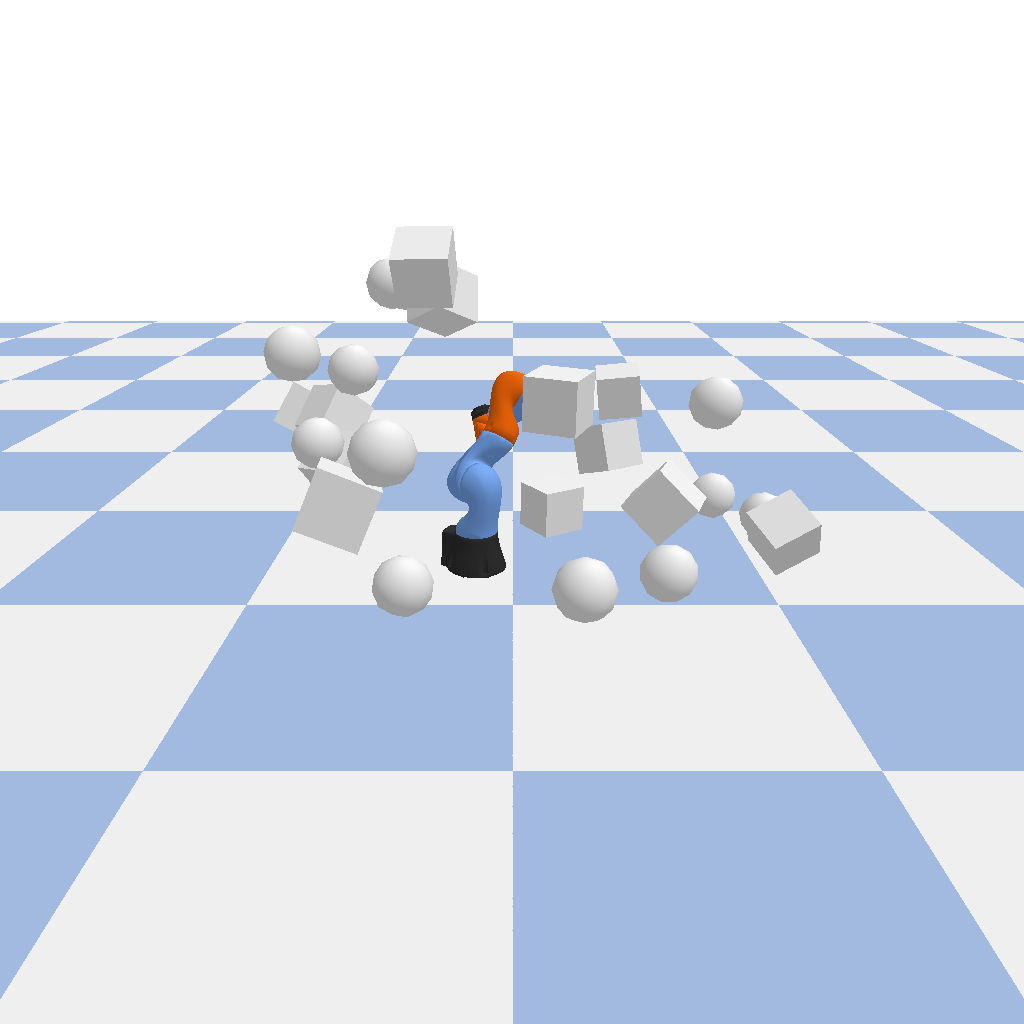}\label{fig:7dof}}
    \hfill
    \subfloat[Two robots.]{\includegraphics[width=0.16\textwidth]{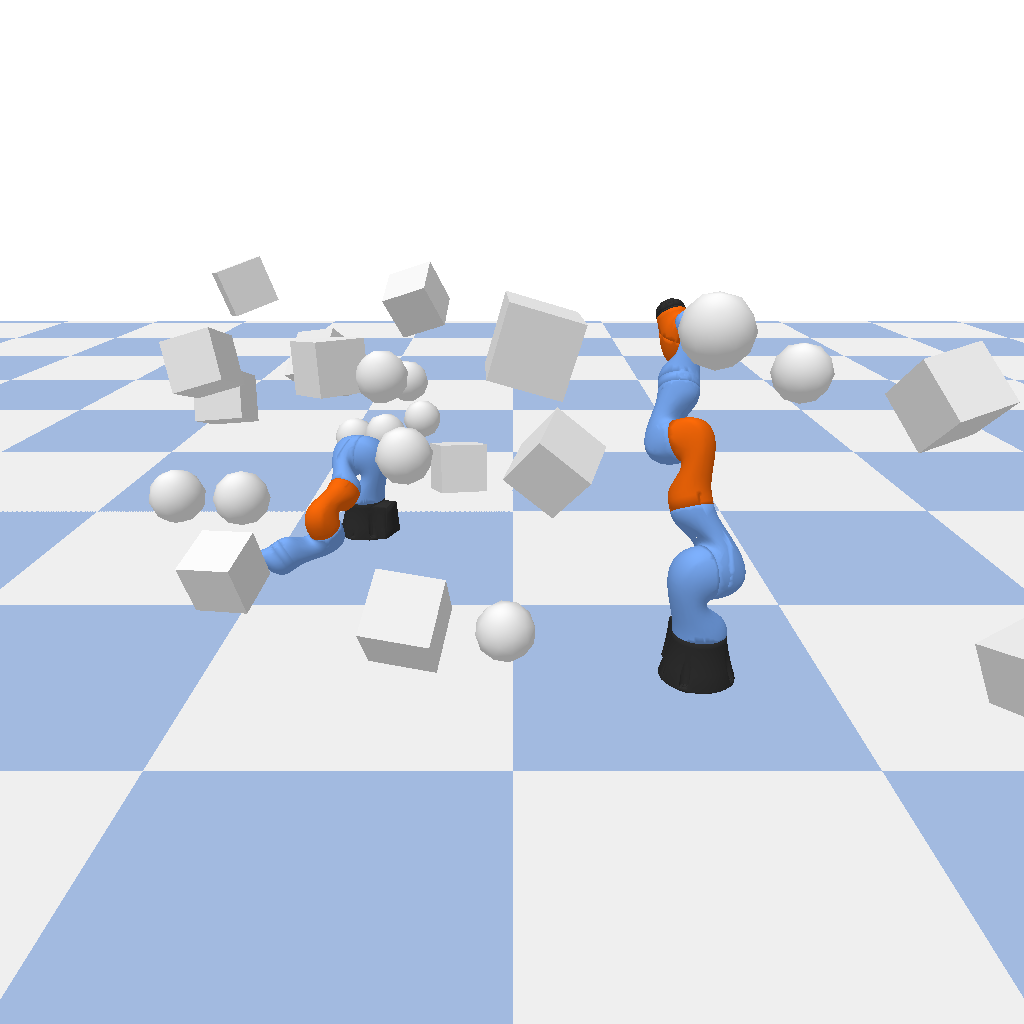}\label{fig:14dof}}
    \hfill
    \subfloat[Three robots.]{\includegraphics[width=0.16\textwidth]{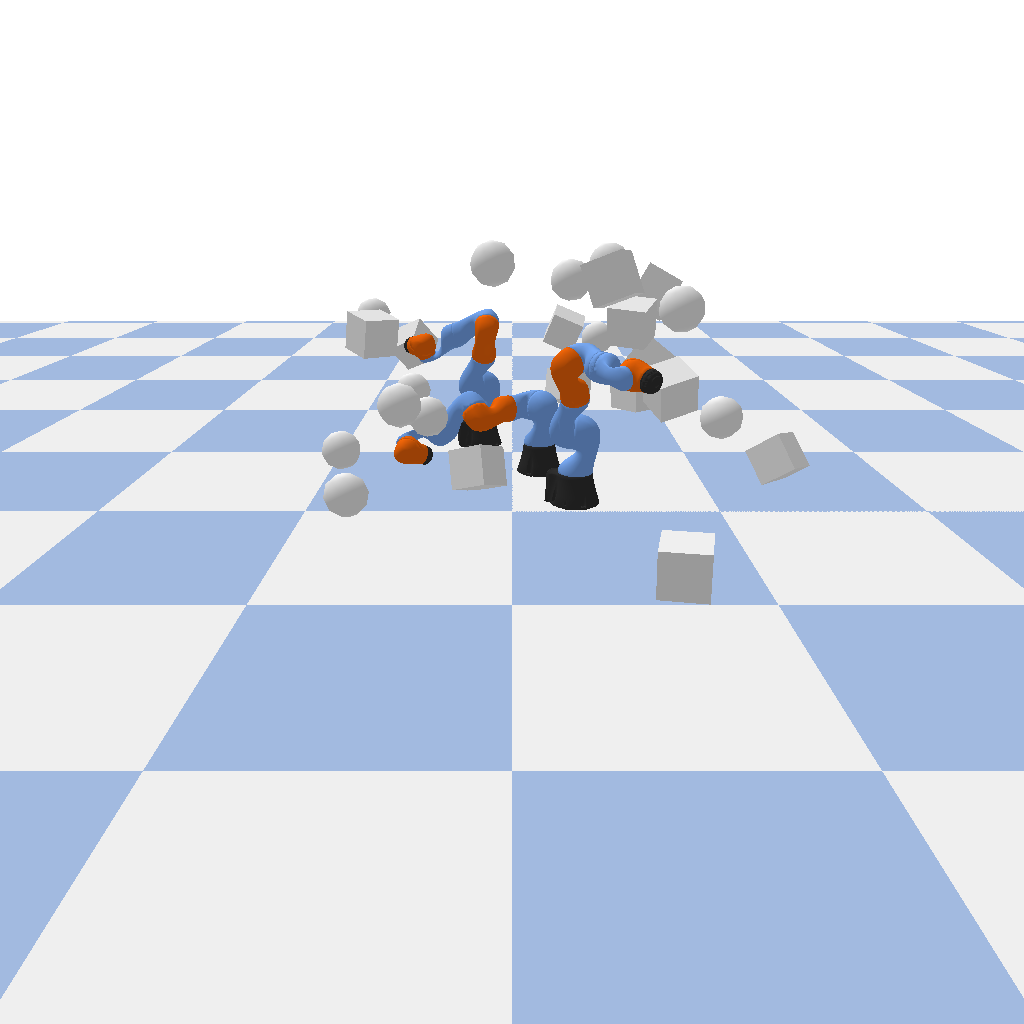}\label{fig:21dof}}
    \hfill
    \subfloat[Four robots.]{\includegraphics[width=0.16\textwidth]{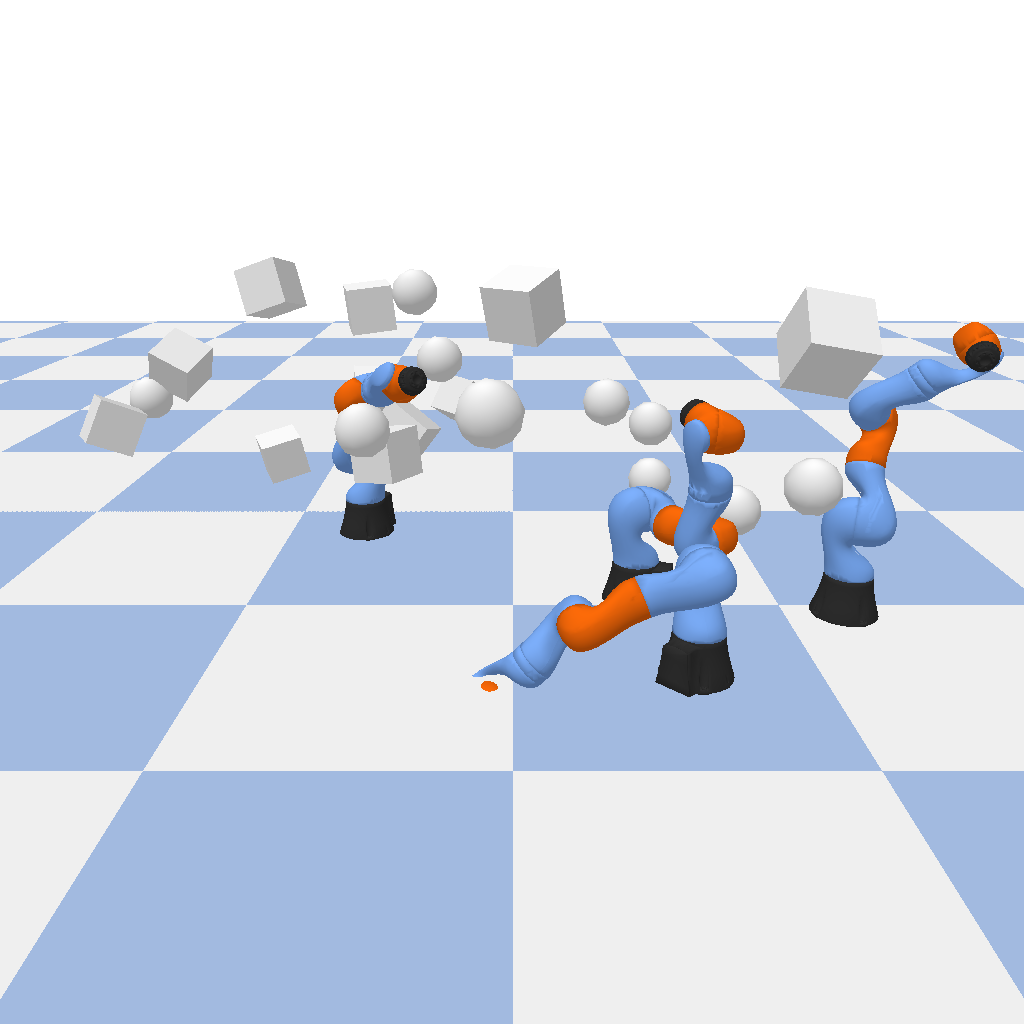}\label{fig:28dof}}
    \hfill
    \subfloat[Five robots.]{\includegraphics[width=0.16\textwidth]{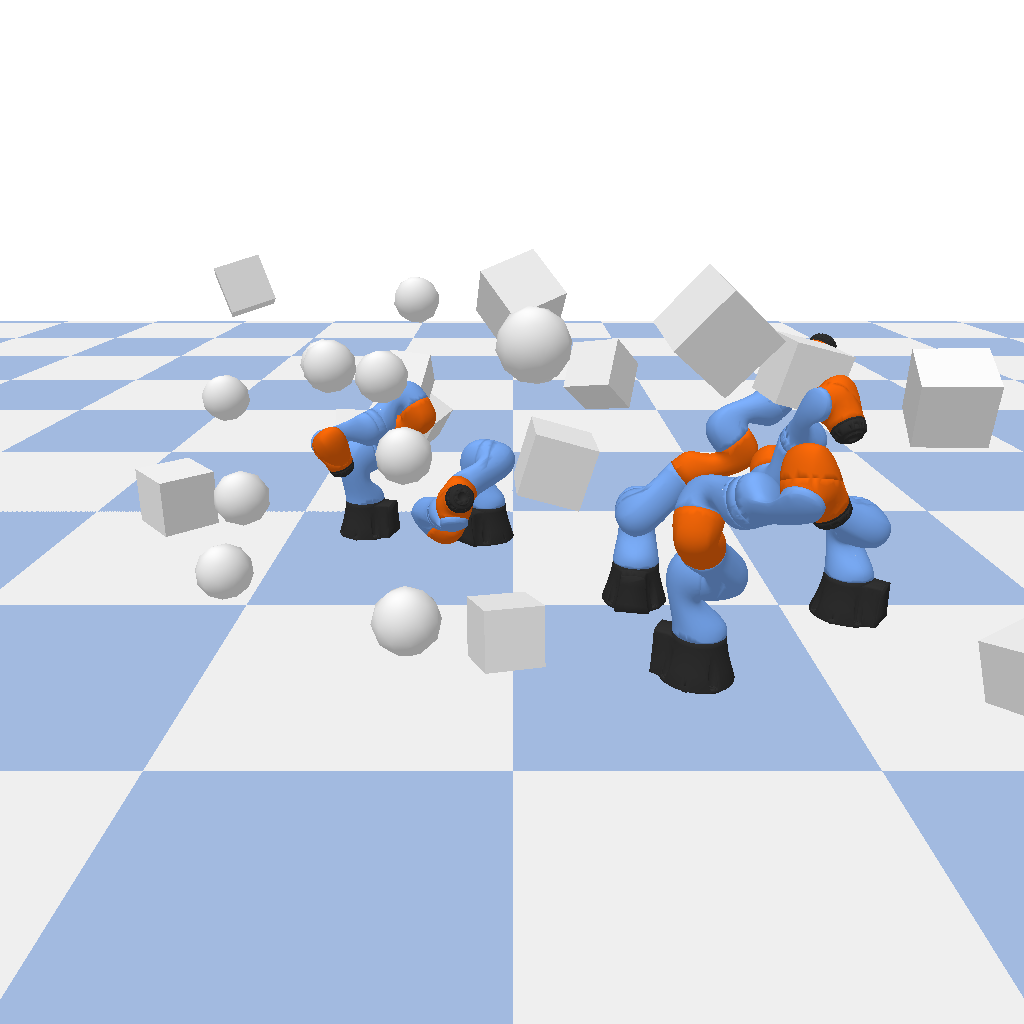}\label{fig:35dof}}
    \hfill
    \subfloat[Six robots.]{\includegraphics[width=0.16\textwidth]{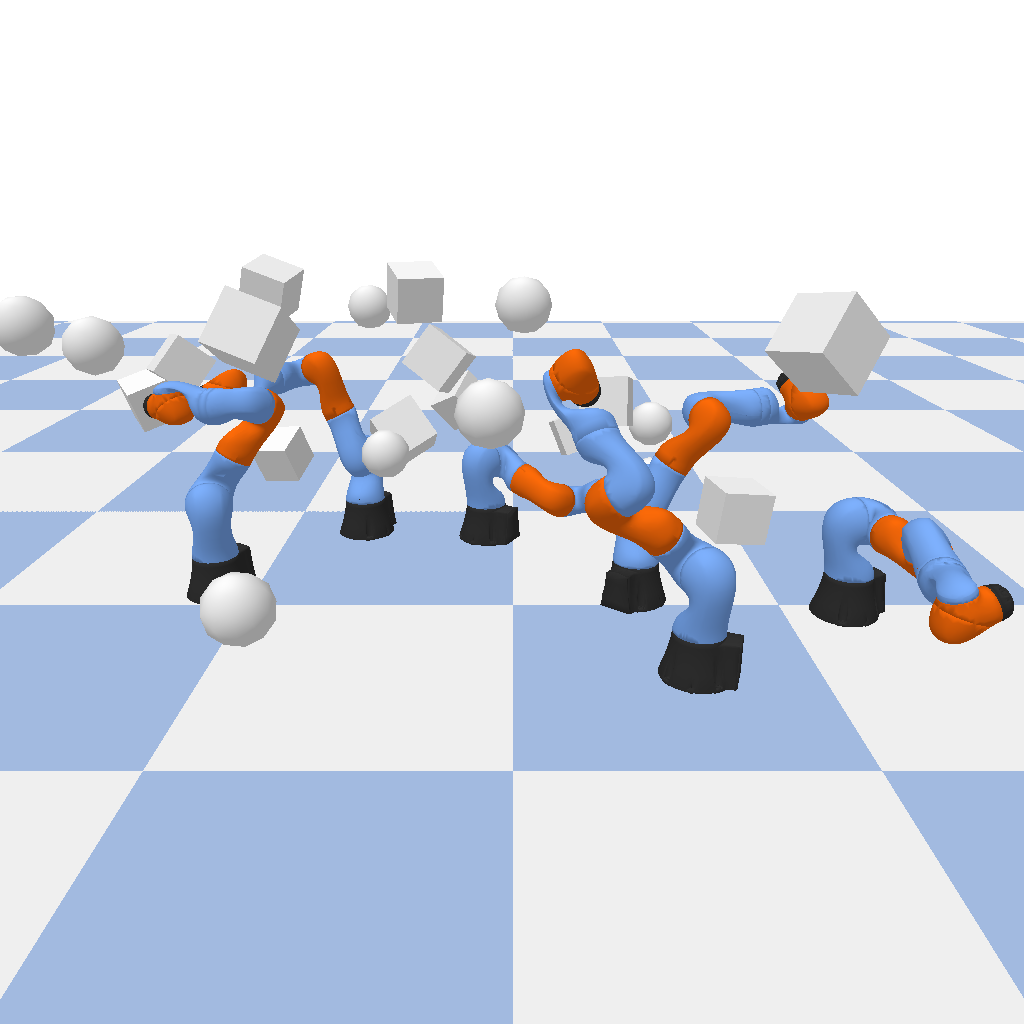}\label{fig:42dof}}
  \caption{\textbf{Environments for Determining Impact of DoF on Performance.} Each environment has randomly placed Kuka LBR iiwa 7 R800 robots, and 25 randomly placed obstacles. Numbers of robots vary from one to six. In the DoF experiments, there are a total of 18 environments ($6 \text{ DoF settings} \times 3 \text{ random seeds per DoF setting}$) like these.}
  \label{fig:dof_envs}
\end{figure}

\begin{figure}
      \subfloat[$10$ obstacles.]
      {\includegraphics[width=0.16\textwidth]{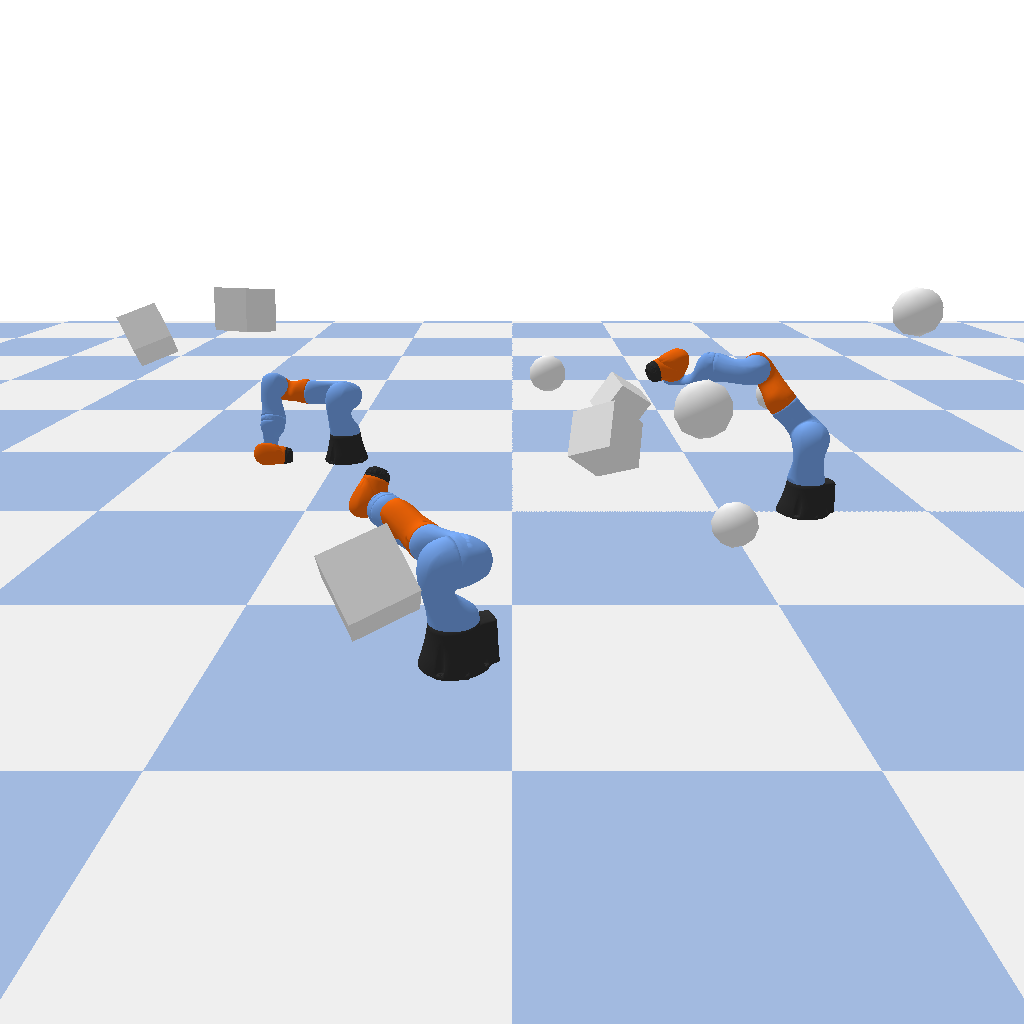}\label{fig:collision_10}}
      \hfill
      \subfloat[$20$ obstacles.]
      {\includegraphics[width=0.16\textwidth]{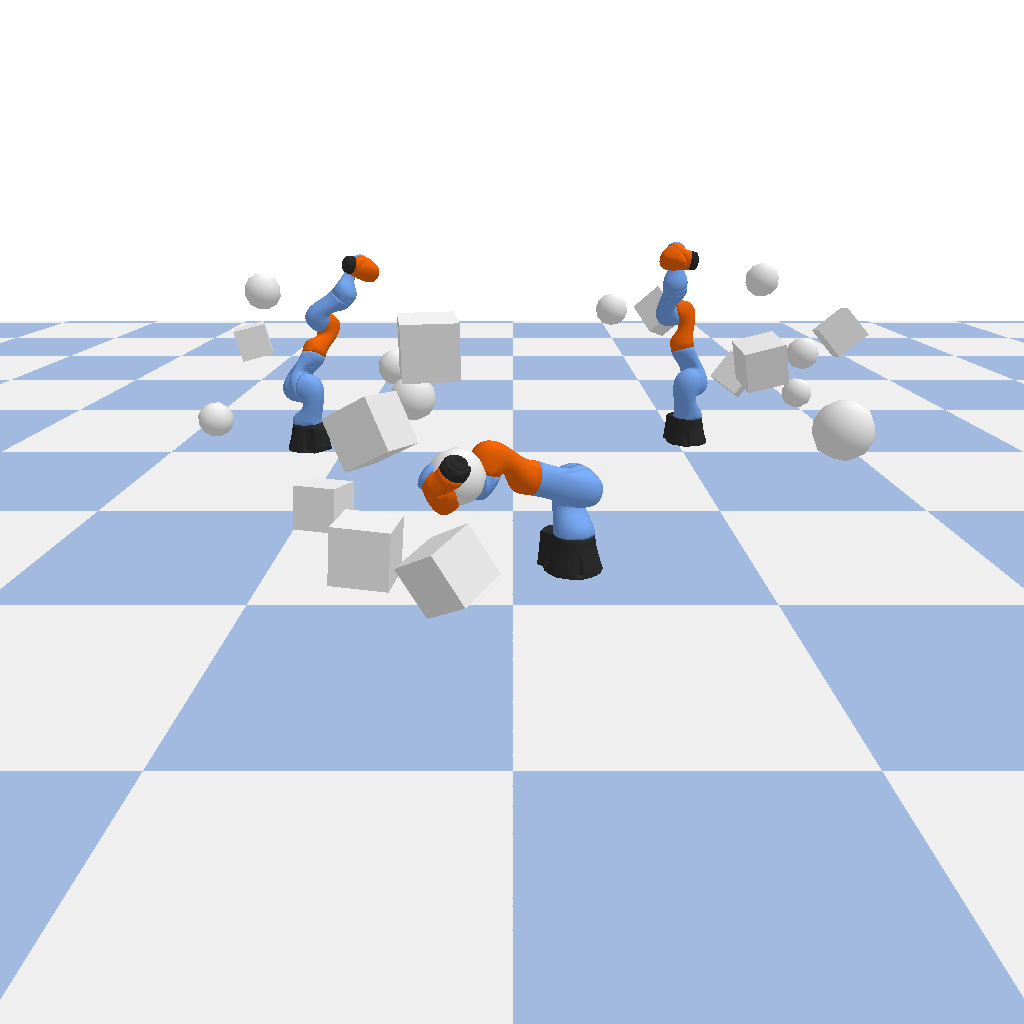}\label{fig:collision_20}}
      \hfill
      \subfloat[$30$ obstacles.]
      {\includegraphics[width=0.16\textwidth]{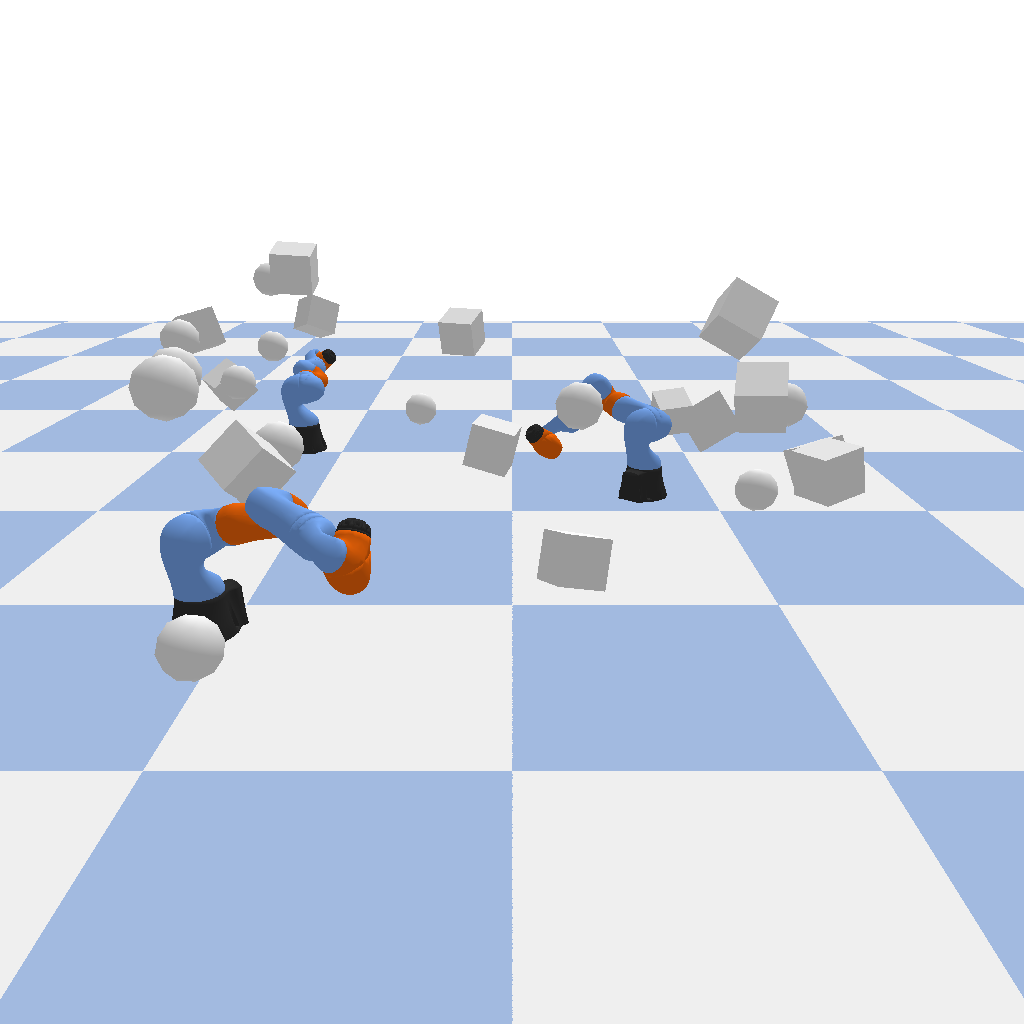}\label{fig:collision_30}}
      \hfill
      \subfloat[$40$ obstacles.]
      {\includegraphics[width=0.16\textwidth]{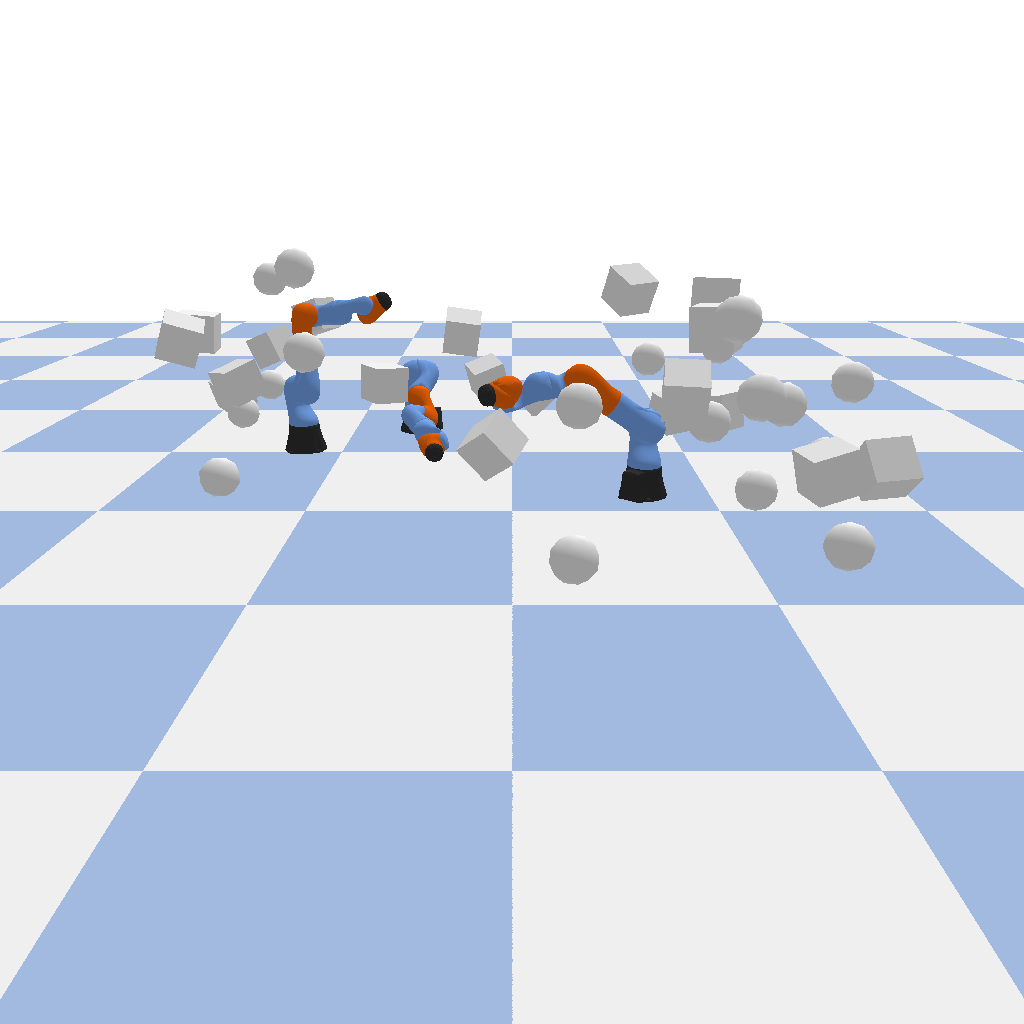}\label{fig:collision_40}}
      \hfill
      \subfloat[$50$ obstacles.]
      {\includegraphics[width=0.16\textwidth]{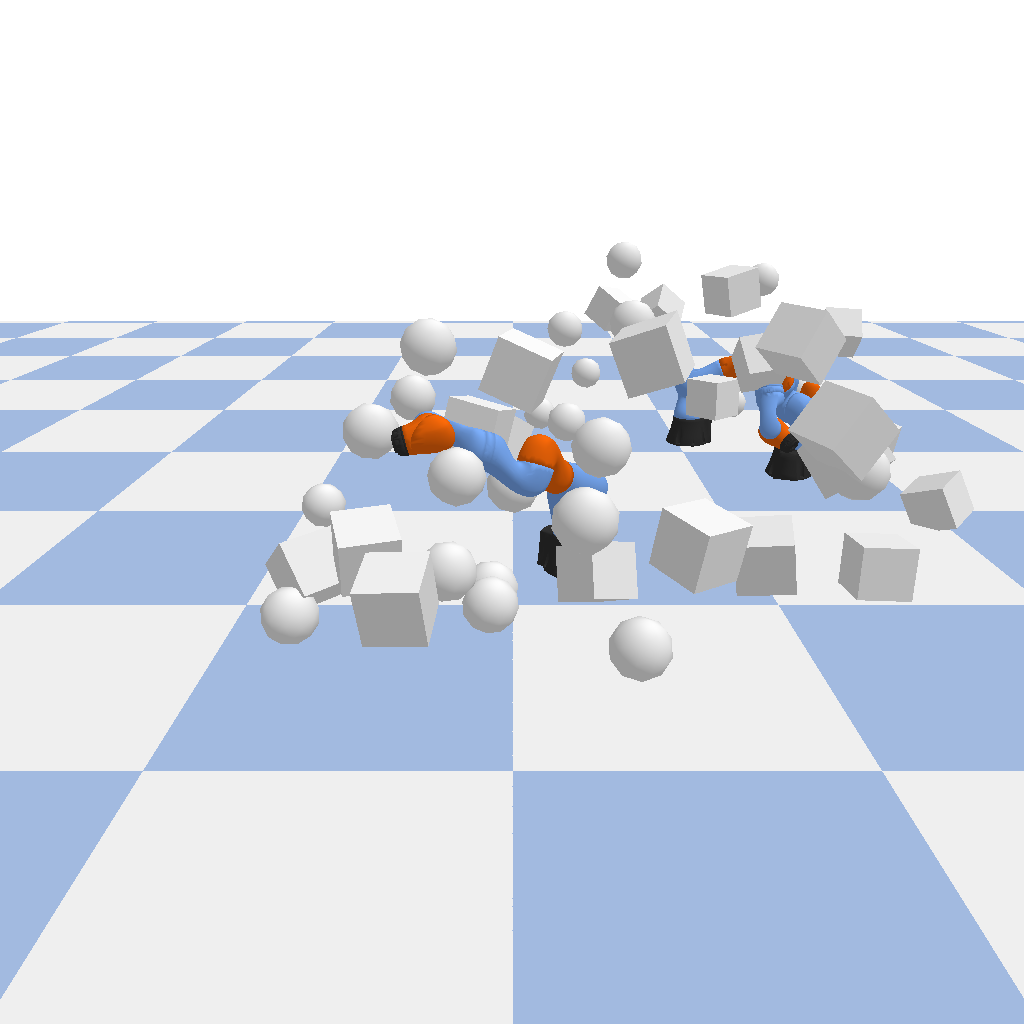}\label{fig:collision_50}}
      \hfill
      \subfloat[$60$ obstacles.]
      {\includegraphics[width=0.16\textwidth]{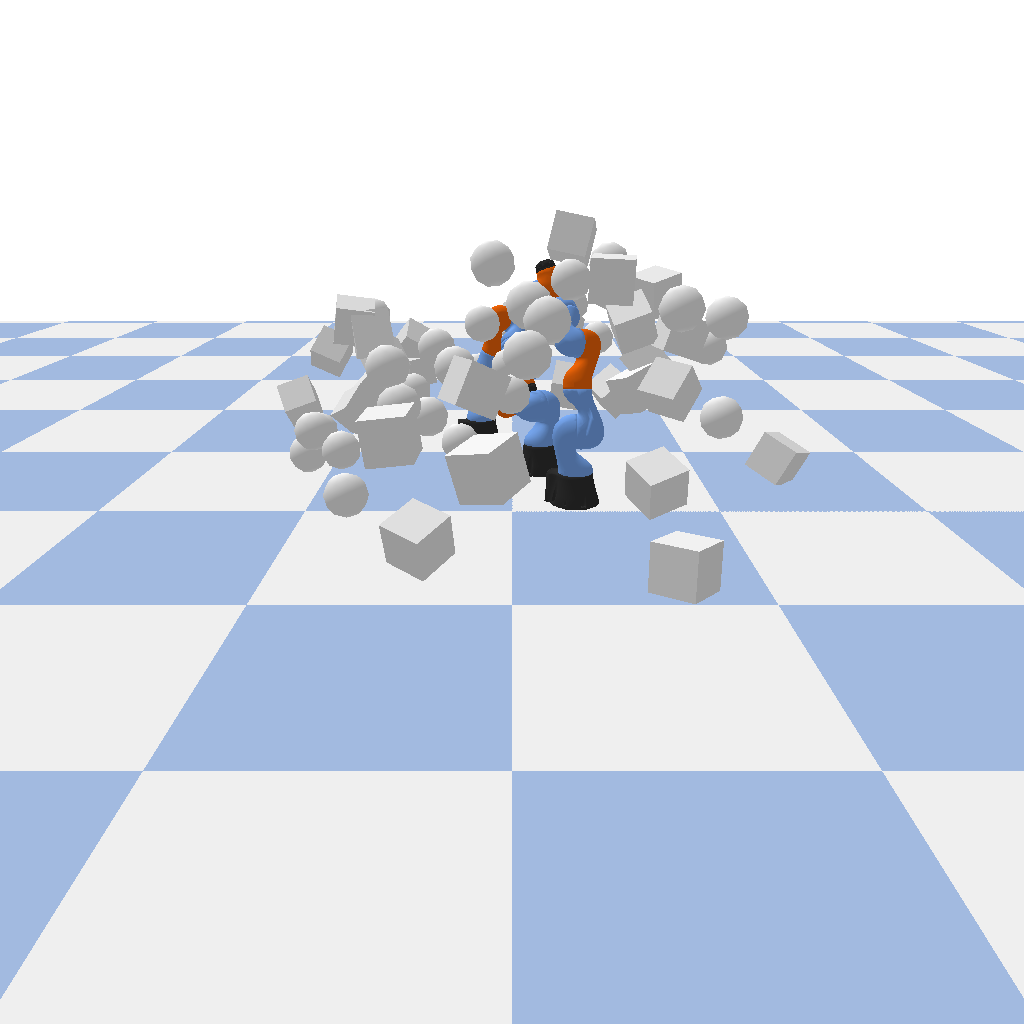}\label{fig:collision_60}}
  \caption{\textbf{Environments for Determining Impact of Collision Density on Performance.} Each environment has three randomly placed Kuka LBR iiwa 7 R800 robots (21 DoF). Number of obstacles varies from $10$ to $60$. In the collision density experiments, there are a total of 36 environments like these, generated with a variety of random seeds.}
  \label{fig:collision_envs}
\end{figure}

%% file: experimental_setup/models_compared.tex
\subsection{Ground Truth Collision Detection}

We use PyBullet for ground-truth collision detection \cite{coumans2016pybullet}. Based on our understanding of the source code on GitHub, PyBullet internally uses the standard GJK \cite{gjk_gilbert1988fast} algorithm for collision detection (along with axis-aligned bounding boxes and EPA \cite{epa_van2001proximity}). 
As stated earlier, we do not explicitly compare our method's speed to the ground-truth collision detection method, because the ground-truth calculates from geometry, while we are trying to learn from samples. (Also, PyBullet has significant computational overhead, which is hard to disentangle from the actual collision computation.)

\subsection{Methods Compared}

\begin{table}[h]
\centering
\begin{tabular}{l|l}
\hline
\rowcolor[gray]{0.95} \textbf{Hyperparameter} & \textbf{Possible Values} \\
\hline
$\mathcal{S}_{max}$ (max \# support points) & $3000, 10000, 30000$ \\
$\mathcal{I}_{max}$ (max \# updates) & $\mathbf{5000}, 30000$ \\
$\gamma$ (kernel width) & $1, \mathbf{5}, 10$ \\
$\beta$ (positive bias) & $1, \mathbf{500}, 1000$ \\
\hline
\end{tabular}
\caption{\textbf{Fastron Hyperparameter Sweep:} We use the recommended hyperparameters (\textbf{bolded}) from \cite{fastron_das2020learning}, and also test a few more to give it a fair chance (we try all $3 \times 2 \times 3 \times 3 = 54$ combinations).}
\label{tab:hyperparameters_fastron}
\end{table}

\begin{table}[h]
\centering
\begin{tabular}{l|l}
\hline
\rowcolor[gray]{0.95} 
\textbf{Hyperparameter} & \textbf{Possible Values} \\
\hline
$|L|$ (\# positional freq.) & 4, 8, 12 \\
\hline
$\beta$ (positive bias) & 1, 2, 5 \\
\hline
$\sigma$ (positional freq. increment) & $\frac{1}{2}$, 1, 2 \\
\hline
\end{tabular}
\caption{\textbf{DeepCollide Hyperparameter Sweep:} The $3 \times 3 \times 3 = 27$ combinations of hyperparameters we tried for our method.}
\label{tab:hyperparameters_dl}
\end{table}

We benchmark our model against Fastron with Forward Kinematics Kernel (Fastron FK), which is known to be state-of-the-art in learning-based collision detection \cite{fastron_das2020learning, forward_kinematics_kernel_das2020forward}.
We note that Fastron and Fastron FK were already benchmarked against the previous state-of-the-art approaches: SSVM \cite{ssvm_huang2010sparse}, ISVM \cite{cspace_svm_pan2015efficient}, and GJK \cite{gjk_gilbert1988fast}. In the extensive experiments presented by the authors, Fastron and Fastron FK were shown to be superior to or on-par with all the aforementioned methods, as it relates to the speed-error tradeoff (albeit at $\leq 7$ DoF, with only a few thousand training samples) \cite{fastron_das2020learning, forward_kinematics_kernel_das2020forward}. Furthermore, they found that Fastron FK is superior to vanilla Fastron, in terms of both speed and time \cite{forward_kinematics_kernel_das2020forward}. Due to these reasons, the theoretical analysis described in Section \ref{sec:theoretical}, and the apparent lack of open-source code for other learning-based methods, we only benchmark against Fastron FK.


\noindent\textbf{Hyperparameters:}
In order to fairly compare the methods, we test multiple hyperparameter settings. We made sure to include the hyperparameter settings included by the authors of Fastron, as those were previously determined to be near-optimal \cite{fastron_das2020learning}. See Tables \ref{tab:hyperparameters_fastron} and \ref{tab:hyperparameters_dl}.

\noindent\textbf{Train-Test Split:}
For almost all of the experiments, we use $30,000$ training points and $5,000$ testing points. All points are uniformly randomly sampled from the configuration space. The only exception is when we investigate the impact of training dataset size on model performance -- here we vary the number of training samples between $10^{2}$ and $10^{5}$, but still use $5,000$ testing samples.



%% file: experimental_setup/metrics.tex
\subsection{Metrics}

In creating a model for collision detection, we wish to optimize on two fronts: speed and correctness.

As it relates to speed, we measure (following \cite{fastron_das2020learning}):
\begin{itemize}
\item \textbf{Training Time:} This is the time it takes to fit our model to the given data on the workspace. 
\item \textbf{Time per Inference:} This measures the average time it takes the model to tell us whether a singular configuration will collide. We measure this separately from training time because the training time is a one-time cost, but after we have the trained model, we can make as many queries as we want.
\end{itemize}
We leave out the forward kinematics kernel computation \cite{forward_kinematics_kernel_das2020forward} from our speed calculations, because all the methods (\textit{i.e.}, DeepCollide, Fastron FK) we compare use it. So, whether or not we include it, the conclusion of which method is fastest remains the same. Furthermore, since we want to find out which method is the fastest, we believe it is more illuminative to isolate and base our comparison on the parts of the methods that are different.

As it relates to correctness, we measure (again, after \cite{fastron_das2020learning}):
\begin{itemize}
    \item \textbf{Accuracy:} This is defined as $\frac{TP + TN}{TP + TN + FP + FN}$, where $TP$ is the number of true positives, $TN$ is the number of true negatives, $FP$ is the number of false positives, and $FN$ is the number of false negatives, from the model's predictions. This is the standard metric used in machine learning applications.
    \item \textbf{TPR:} This is defined as $\frac{TP}{TP + FN}$. Intuitively, it quantifies the percent of actual collisions that our model catches. A high value for this metric indicates that our model will be able to show the robot where to avoid obstacles.
    \item \textbf{TNR:} This is defined as $\frac{TN}{TN + FP}$. We can think of this as the percent of the free space that our model correctly identifies. This metric is important because it shows how good the collision detection function will be at showing the robot collision-free paths to its destination.
\end{itemize}

%% file: results.tex
\input{results/model.tex}

\input{results/dof.tex}

\input{results/collision_density.tex}

\input{results/sample_size.tex}


%% file: results/model.tex
\section{Impact of Model}


\begin{figure}
  \centering
  \subfloat[Error (1 - Accuracy) vs. Inference Time
  ]{\includegraphics[width=0.45\textwidth]{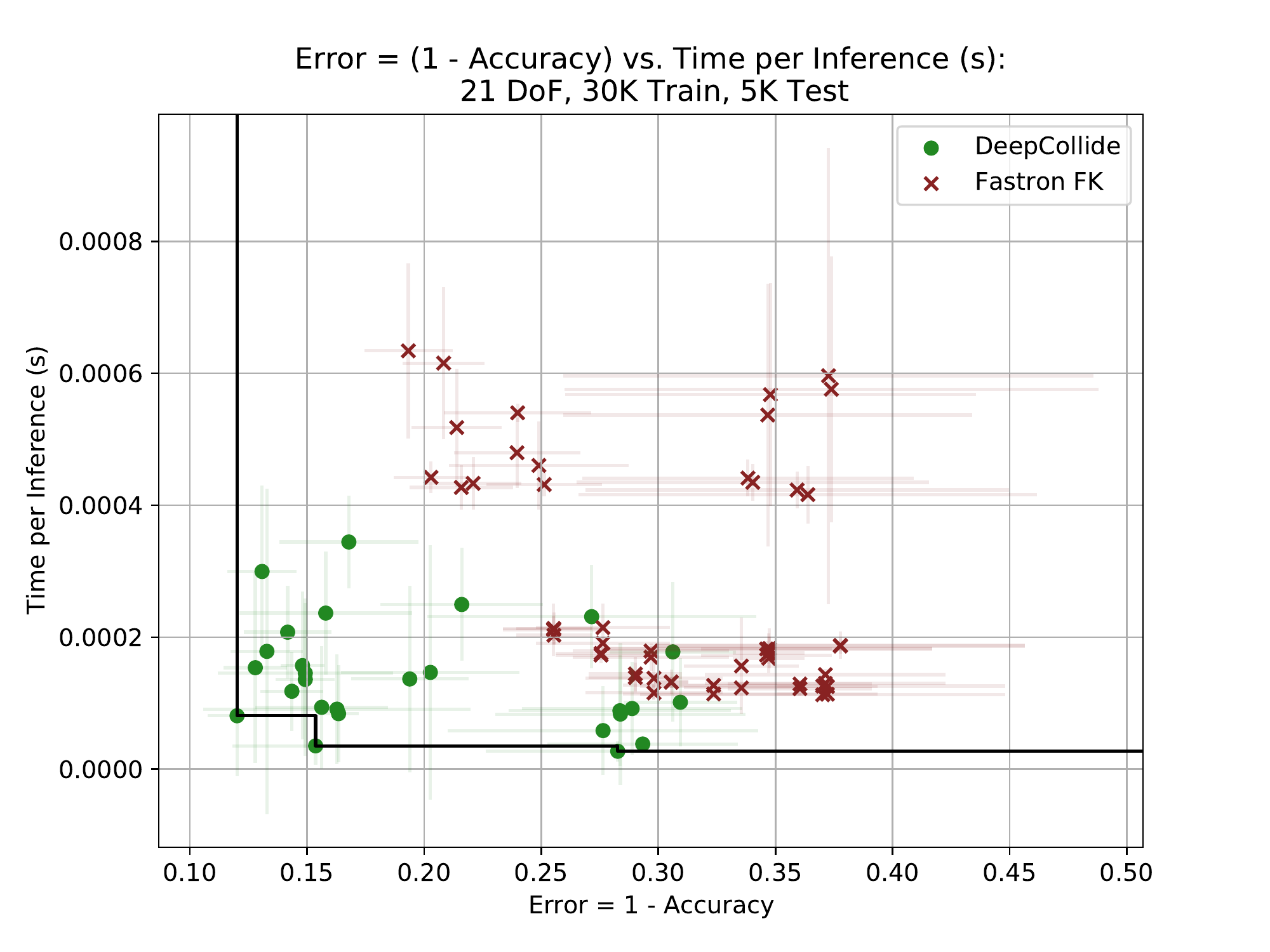}\label{fig:acc_v_inferenceTime}}
  \hfill
  \subfloat[Error (1 - Accuracy) vs. Train Time
  ]{\includegraphics[width=0.45\textwidth]{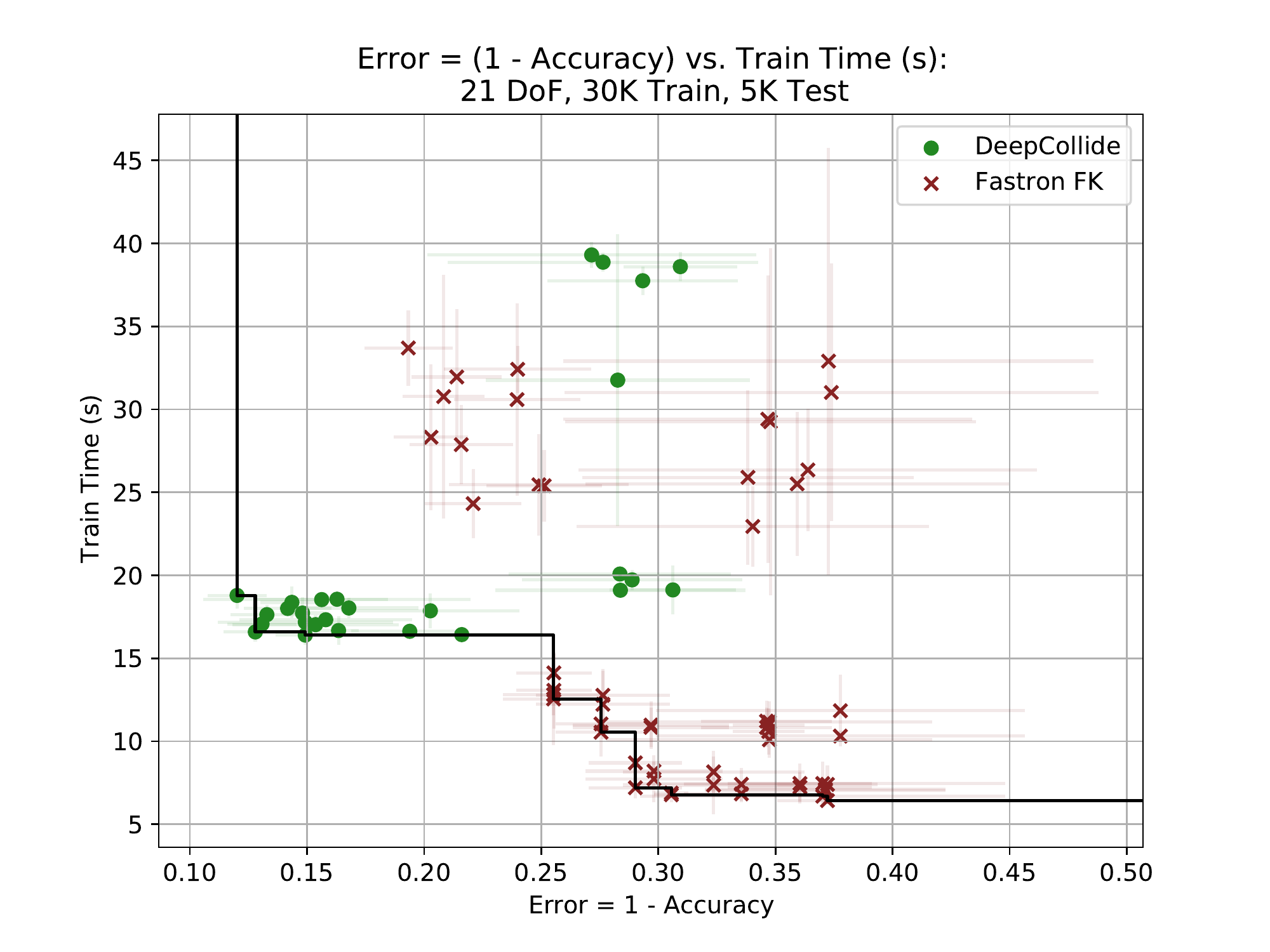}\label{fig:acc_v_trainTime}}
  \hfill
  \subfloat[TPR error vs. TNR error]{\includegraphics[width=0.45\textwidth]{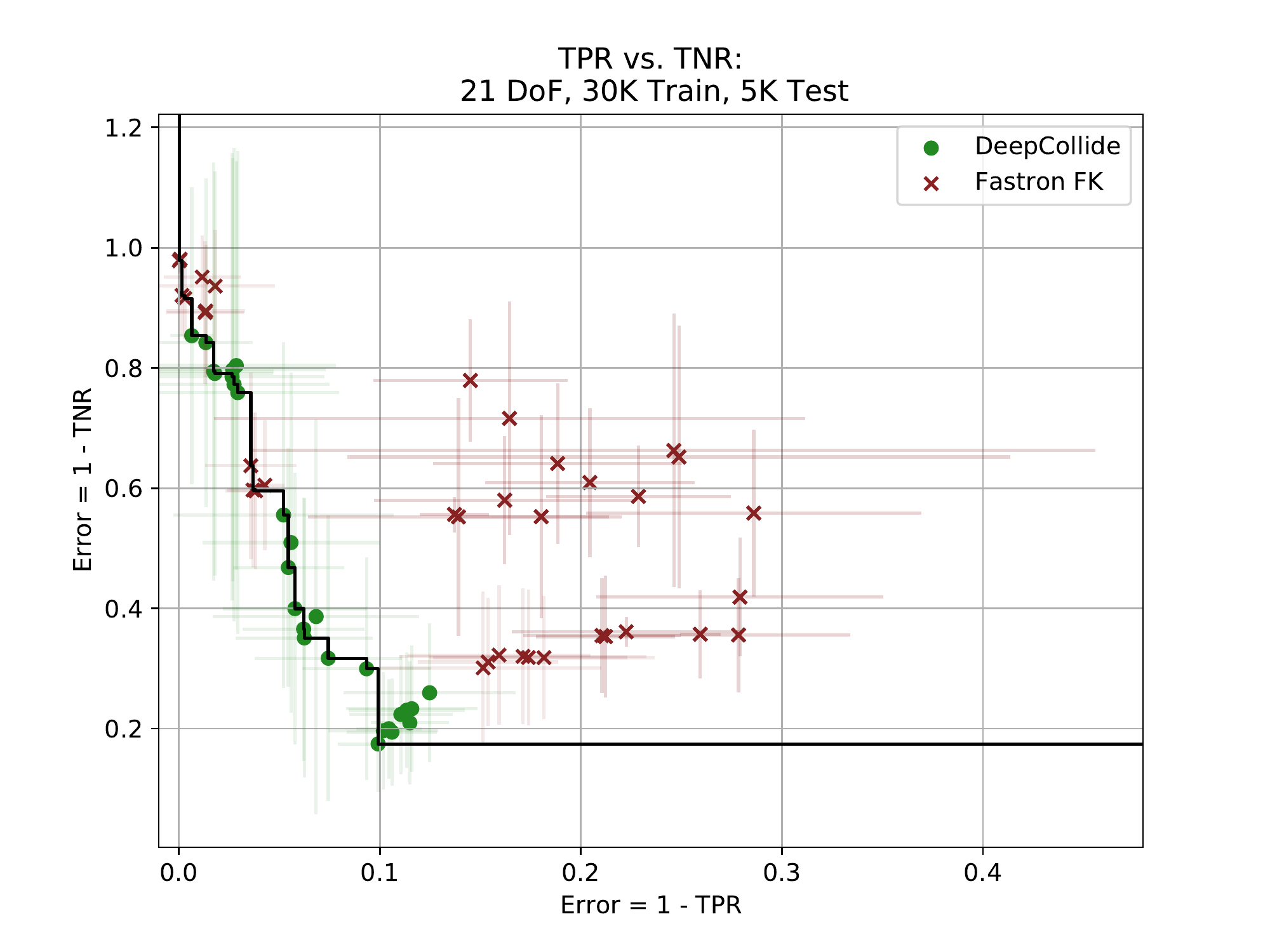}\label{fig:tpr_v_tnr}}
  \hfill
  \caption{\textbf{DeepCollide vs. Fastron FK (SoTA) for collision detection.} In Figures \ref{fig:acc_v_inferenceTime}, \ref{fig:tpr_v_tnr}, and \ref{fig:acc_v_trainTime}; each mark (green dots for DeepCollide, red crosses for Fastron) represents a version of the model with different hyperparameters; error bars represent the standard deviation of results, over the three randomly generated testing environments (depicted in Figures \ref{fig:pareto_env0}, \ref{fig:pareto_env1}, and \ref{fig:pareto_env2}).}
  \label{fig:pareto}
\end{figure}

See Figure \ref{fig:pareto}, which shows the Pareto frontiers with respect to the inference time vs. accuracy tradeoff, and the true positive rate vs. true negative rate tradeoff. 
The marks show the performance of Fastron FK and DeepCollide at various hyperparameter settings, while the error bars represent the standard deviation of the model versions over the three randomly generated testing environments. 

\subsection{Speed vs. Correctness Trade-off}

We choose inference time vs. error as defined by accuracy for Figure \ref{fig:acc_v_inferenceTime}'s Pareto frontier, because the goal of this work is to provide the optimal balance between speed and correctness. As for why we specifically choose inference time as our speed metric in this Pareto frontier, asymptotically, speed can be expressed as a function of inference time alone. We choose accuracy as our correctness metric because because it accounts for all of the categories in the confusion matrix (true positives, true negatives, false positives, false negatives); while other metrics (like TPR) do not account for some of the categories. 

We see that DeepCollide makes up all of the points on the Pareto-optimal frontier, so we can conclude that DeepCollide outperforms Fastron FK in regards to the speed-correctness trade-off. We can be confident in this conclusion, because we did it over a variety of hyperparameters, including the hyperparameters that the authors of Fastron deemed to be (near-)optimal for their model \cite{fastron_das2020learning}. Furthermore, we conducted these evaluations over multiple random environments with varying collision densities, so we are fairly confident that the result is not environment-specific.

Furthermore, while training time does not matter in the asymptotic case (predicting $n \rightarrow \infty$, where $n$ is the number of points predicted), it still is important for practical applications. Thus, we also investigate the trade-off between train time and accuracy in Figure \ref{fig:acc_v_trainTime}. 
Here, both DeepCollide and Fastron FK have instances on the Pareto-optimal frontier. We observe that Fastron FK has the fastest overall models, at the cost of accuracy, while DeepCollide has the most accurate overall models, at the cost of speed. This is emblematic of the general differences between neural networks and traditional machine learning methods -- neural networks are typically more accurate, but often require extra training time to achieve their high accuracy. 

A natural question that may arise is how we are able to achieve faster inference times than and comparable train times to Fastron, when our DeepCollide neural network has hundreds of thousands of parameters. 
Actually, this is in line with our theoretical analysis from Section \ref{sec:theoretical}. 
In high-DoF settings, we need tens of thousands of training samples to get an accurate collision detection function, due to the curse of dimensionality. 
Given that Fastron's time per inference is linear with respect to both training set size and DoF, while its training time is nearly quadratic with respect to training set size and linear with respect to DoF; this quickly becomes computationally prohibitive, and essentially limits Fastron to low-DoF cases.
In contrast, even though DeepCollide also has (nominally, as will be shown in Section \ref{sec:dof_impact_empirical}) linear time complexity with respect to DoF, its time per inference has no dependency on training dataset size, and its training time is only linear with respect to training dataset size.
Thus, in even a moderate-data case like this evaluation, we see DeepCollide's benefits.

Furthermore, the use of GPUs in training DeepCollide (which the official implementation of Fastron does not support) boosts DeepCollide's training speed. We emphasize that we use CPUs in testing \textit{both} Fastron and Deep Collide. 

\subsection{Error Modality Trade-off}

We choose error as defined by TPR vs. error as defined by TNR for Figure \ref{fig:tpr_v_tnr}'s Pareto frontier, because while accuracy is a good "catch-all" correctness metric, it misses the fine-grained detail of what kinds of errors are made. In particular, we care about two kinds of errors: (1) failing to detect a collision (which can be expressed by $1 - TPR$); (2) failing to find a part of the free path (which can be expressed by $1 - TNR$). Thus, this Pareto chart of TPR error vs. TNR error dives deeper into the error modalities of the methods.

Here, the Pareto-optimal frontier is made up of a combination of DeepCollide and Fastron FK instances, but it is still mostly DeepCollide. Thus, we can say that both models have comparable ability in balancing collision and free path detections, although DeepCollide appears to be slightly better.


\subsection{Hyperparameter Recommendations for DeepCollide}

A hyperparameter setting for DeepCollide that was on the Pareto-optimal frontier for all charts (\textit{i.e.}, accuracy vs. inference time, accuracy vs. train time, TPR vs. TNR) in Figure \ref{fig:pareto} was $|L| = 12, \beta = 1, \sigma = 1$. This corresponds to $88.0\% \pm 1.3\%$ accuracy, $(8.1 \pm 9.2) * 10^{-5}$ seconds per inference, $18.8 \pm 0.8$ seconds to train, $90.1\% \pm 2.0\%$ TPR, and $82.5\% \pm 8.1\%$ TNR, where the mean and standard deviation are calculated across all three environments (\textit{i.e.}, $n = 3$).

%% file: results/dof.tex
\begin{figure}[]
    \centering
    \subfloat[DoF vs. TPR]{\includegraphics[width=0.8\linewidth]{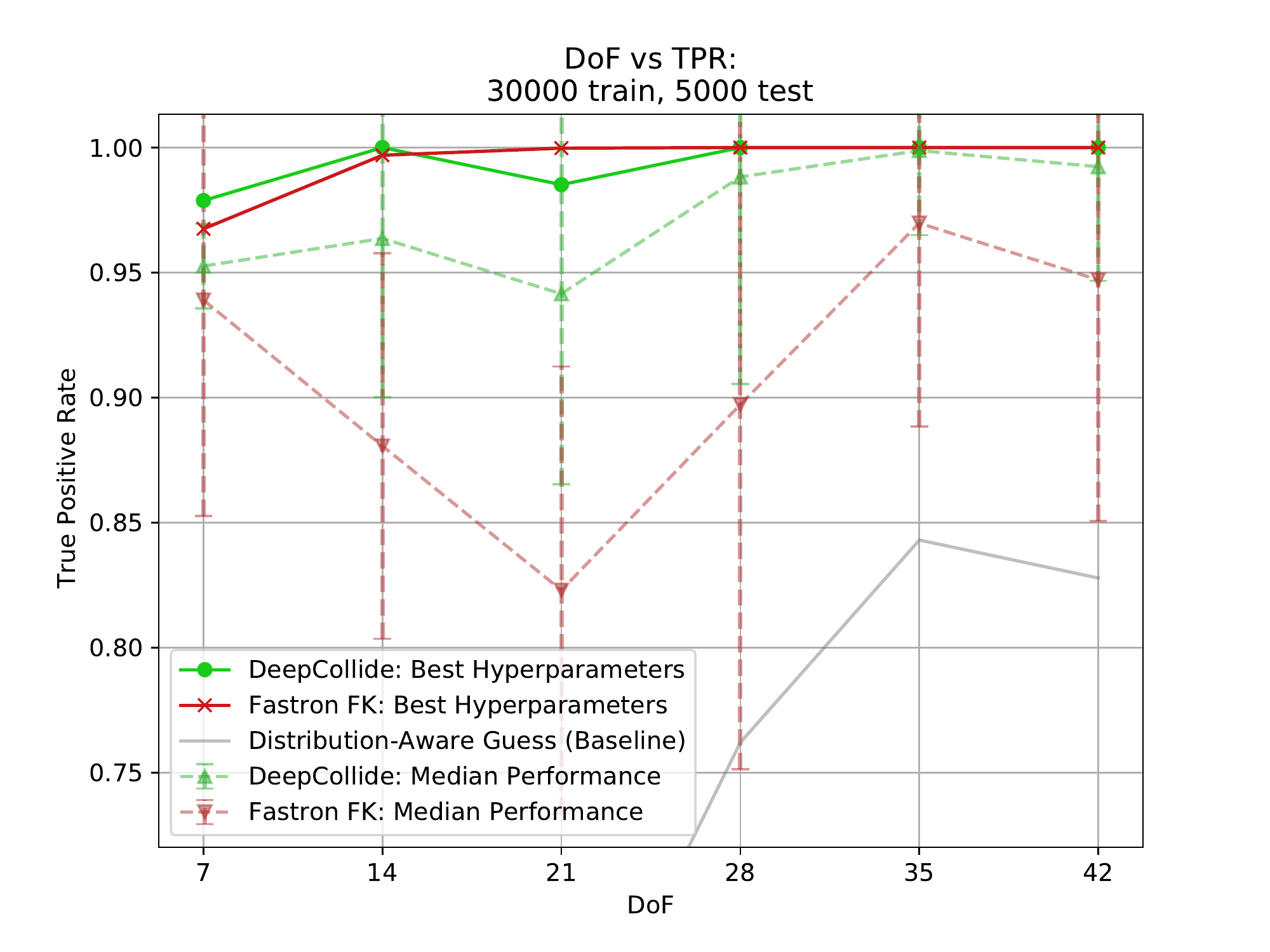}\label{fig:dof_tpr}}
    \hfill
    \subfloat[DoF vs. TNR]{\includegraphics[width=0.8\linewidth]{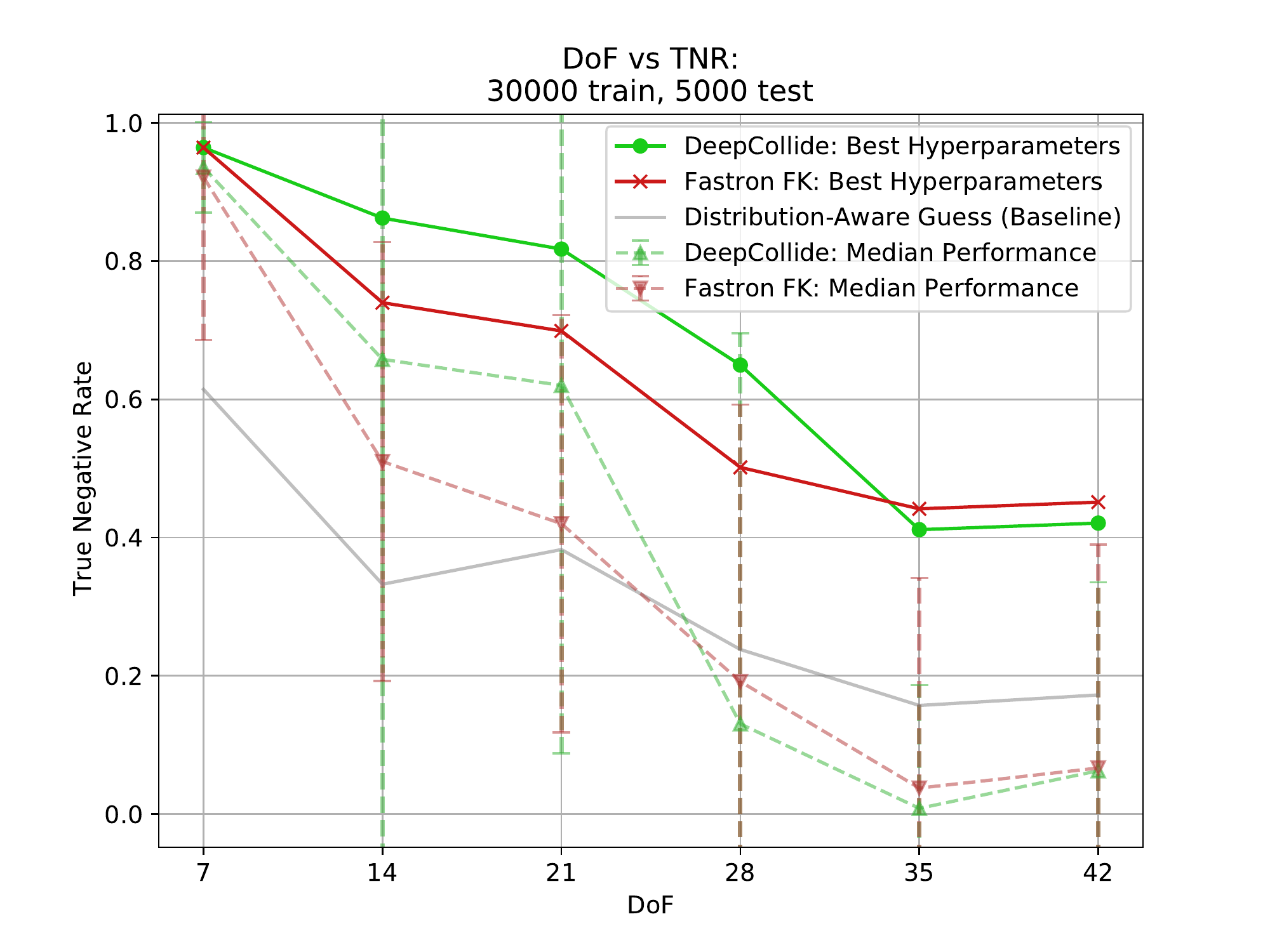}\label{fig:dof_tnr}}
    \hfill
    \subfloat[DoF vs. Accuracy]{\includegraphics[width=0.8\linewidth]{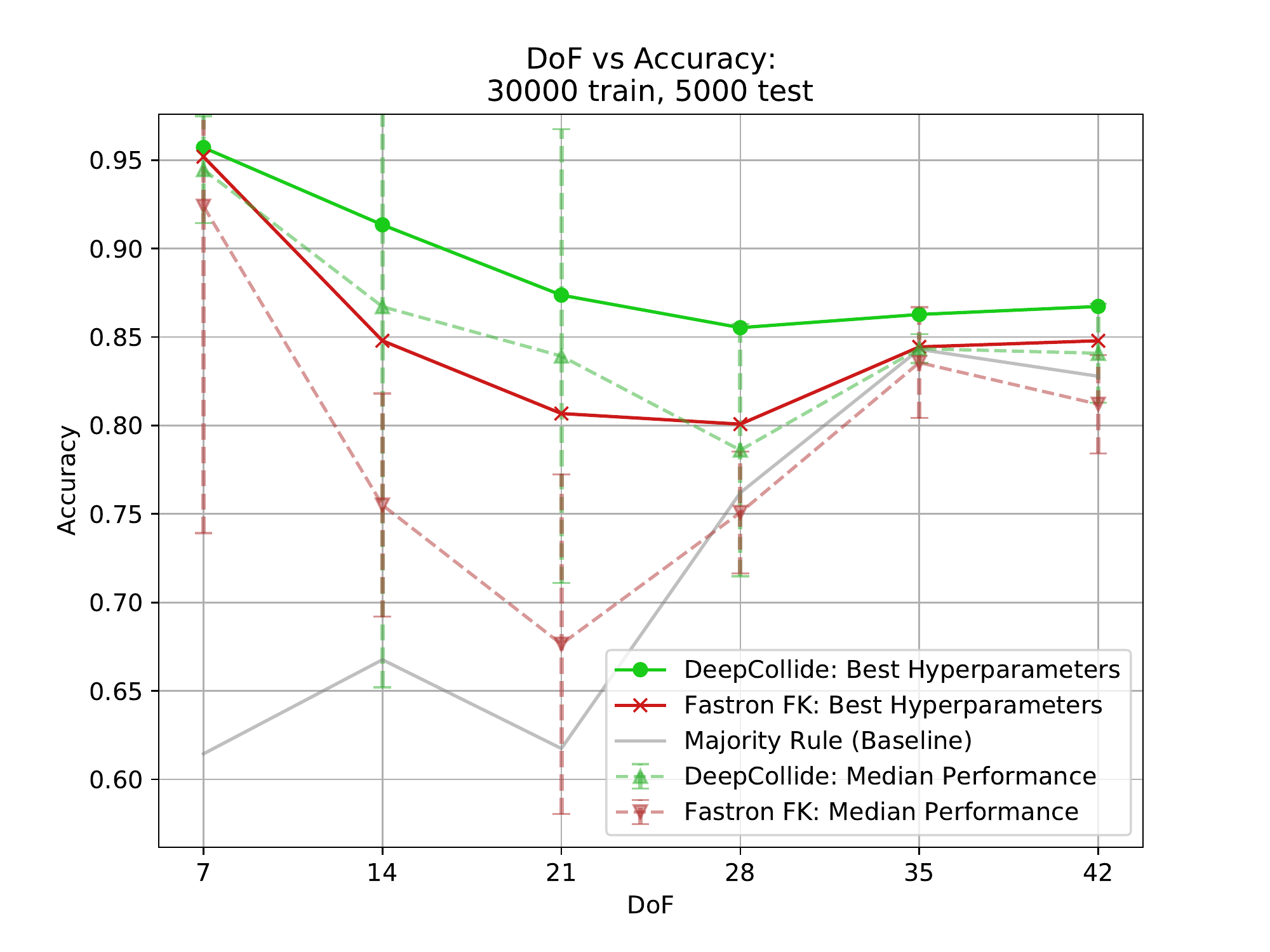}\label{fig:dof_accuracy}}
    \hfill
    \caption{\textbf{Impact of DoF on correctness.} Points on the graph show the performance (averaged over all environments) with the given DoF. Solid lines represent the best-case performance (version of the model with most favorable hyperparameters at that DoF, averaged over all environments) of each model. Dashed lines represent the median performance across all versions (\textit{i.e.}, hyperparameter selection) of the model tested in Tables \ref{tab:hyperparameters_fastron} and \ref{tab:hyperparameters_dl} respectively, with the error bars representing the interquartile range considering all versions of the model. Gray lines represent the dummy classifier performance (for Figure \ref{fig:dof_accuracy}, the strategy is to always select the majority label; for Figures \ref{fig:dof_tpr} and \ref{fig:dof_tnr}, the strategy is to randomly guess based on the percentage of samples that are collisions). Corresponds to environments in Figure \ref{fig:dof_envs}.}
    \label{fig:dof_correctness_impact}
\end{figure}

\begin{figure}[h!]
    \centering
    \subfloat[DoF vs. Train Time]{\includegraphics[width=0.8\linewidth]{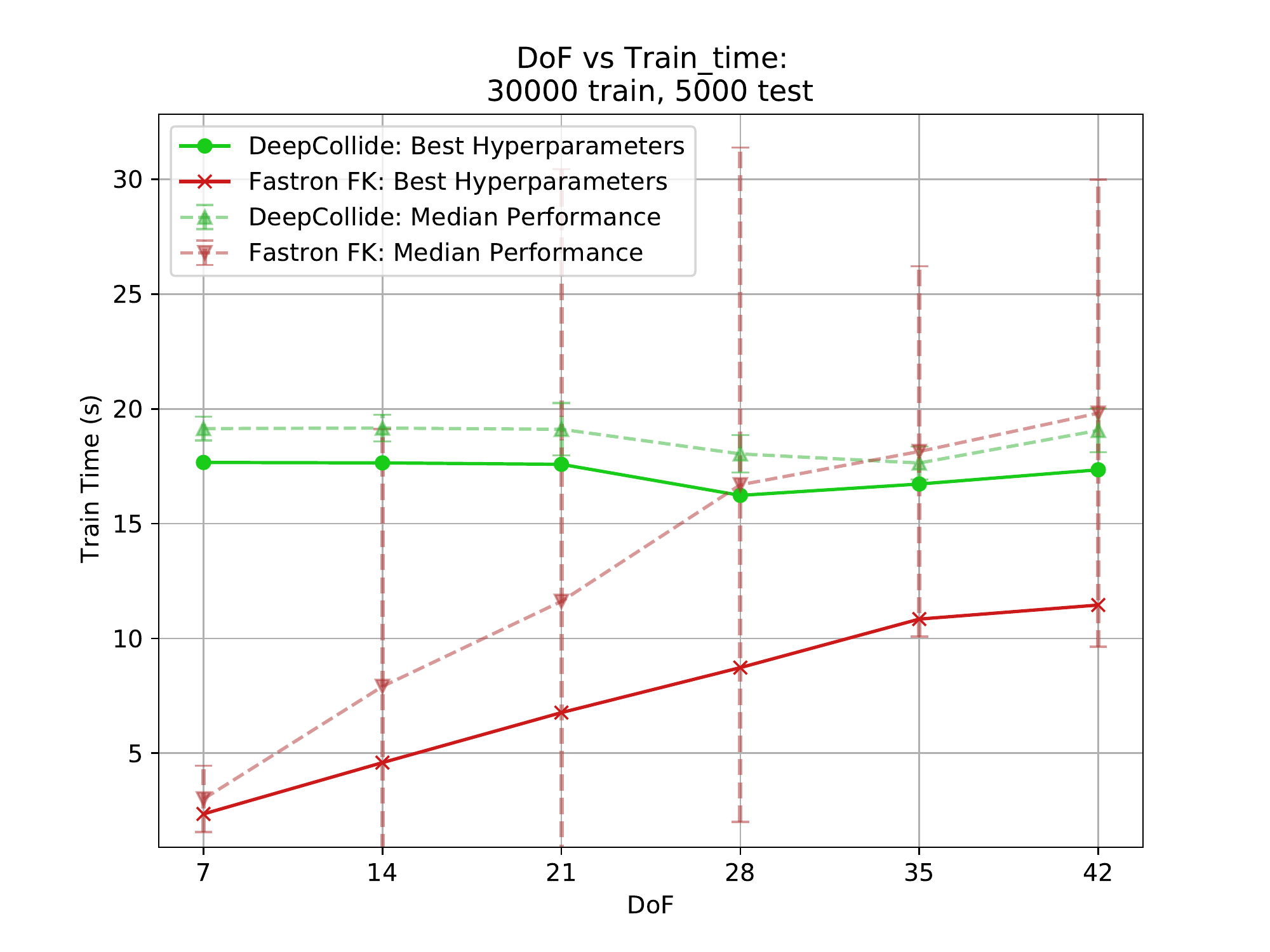}\label{fig:dof_traintime}}
    \hfill
    \subfloat[DoF vs. Test Time]{\includegraphics[width=0.8\linewidth]{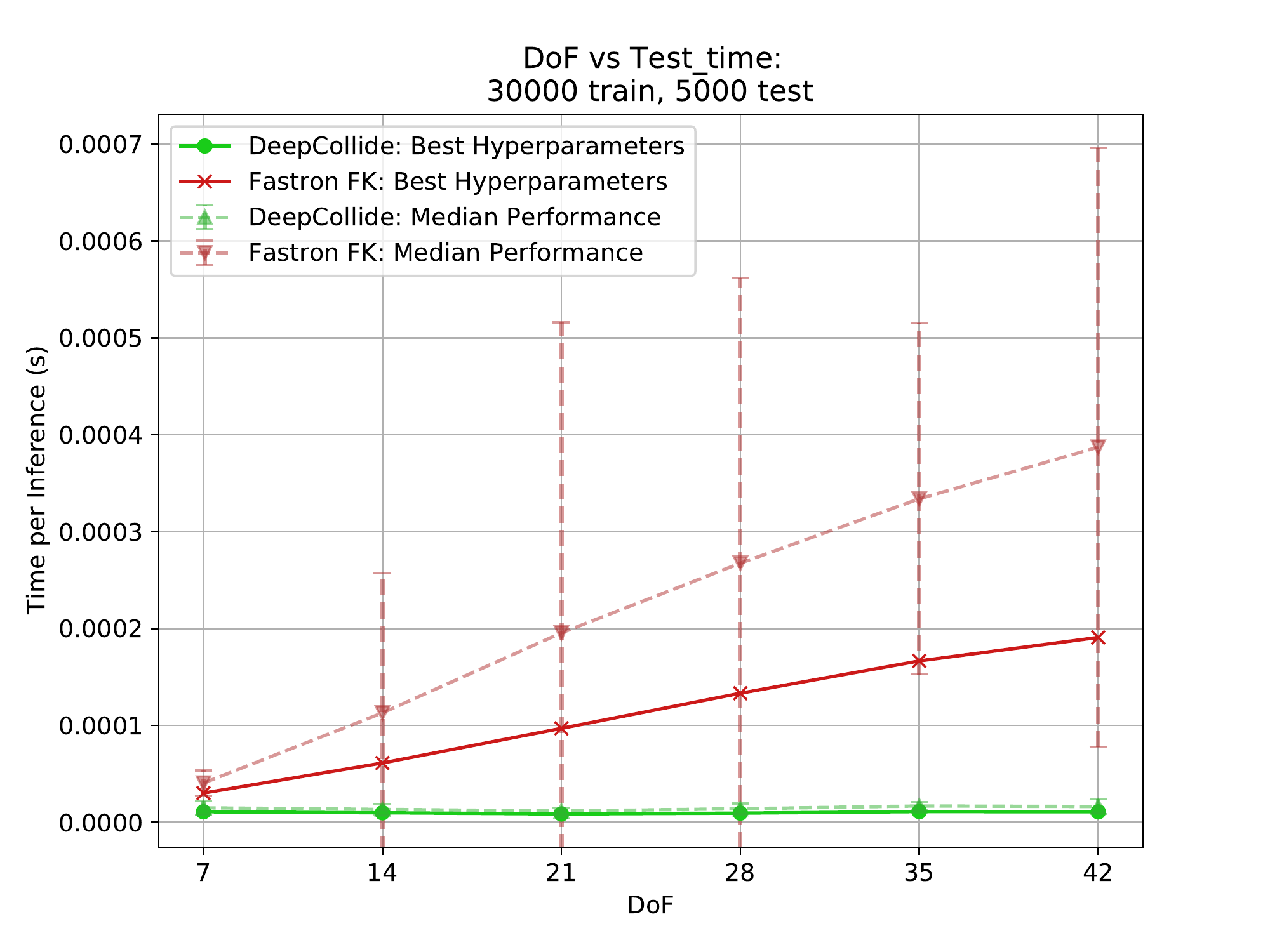}\label{fig:dof_testtime}}
    \hfill
    \caption{\textbf{Impact of DoF on time.} Interpretation is the same as Figure \ref{fig:dof_correctness_impact}.
    }
    \label{fig:dof_time_impact}
\end{figure}

\section{Impact of DoF}
\label{sec:dof_impact_empirical}

\subsection{Impact on Correctness}
Figure \ref{fig:dof_correctness_impact} shows the impact of DoF on model correctness. We see from Figure \ref{fig:dof_accuracy} that DeepCollide has a higher accuracy (averaged over multiple random environments) than Fastron FK at all DoF. We can also see from Figure \ref{fig:dof_tnr} that the true negative rate of DeepCollide is higher than Fastron FK's at lower DoF, and is comparable at higher DoF. Regarding true positive rate, when hyperparameters are tuned well, DeepCollide and Fastron FK run almost equal, but if hyperparameters are not tuned carefully (see dotted median performance lines), DeepCollide is clearly better \footnote{The astute reader will notice that DeepCollide has higher best-case (as it relates to hyperparameter selection) accuracy than Fastron FK at 42 DoF, despite its best-case TNR and TPR being lower than Fastron FK's. The reason for this is that the best-case hyperparameters are not necessarily the same across metrics. A classic example of this, as noted by the authors of Fastron, is that a high bias term $\beta$ will lead to high TPR, but low TNR \cite{fastron_das2020learning}.}.


\subsection{Impact on Speed}

Figure \ref{fig:dof_time_impact} shows the impact of DoF on speed. 
We notice that in train and test time, Fastron's cost roughly scales linearly with the number of DoF. 
In contrast, DeepCollide's time cost is almost constant with respect to DoF.
This is actually faster than the theoretical calculations from Section \ref{sec:theoretical}. These results are surprising, but we verified them on another machine, and came to the same conclusion \footnote{The second machine had AMD Ryzen Threadripper 3960X 24-Core Processor for the CPU, and NVIDIA GeForce RTX 3090s for the GPUs. Runtimes were generally faster for both models on the second machine, but the conclusions of the comparisons were still the same as on our original machine. We noticed a very small positive slope for DeepCollide testing time on the second machine, but it was negligible in comparison to Fastron's testing time increases.}. We think the near-constant runtime is because PyTorch's implementations of tensor operations are highly optimized \cite{paszke2019pytorch}.
However, even with the linear time complexity of Fastron with DoF, DeepCollide is still slower in train time.

%% file: results/collision_density.tex
\begin{figure}[h!]
    \centering
    \subfloat[Collision Density vs. TPR]{\includegraphics[width=0.8\linewidth]{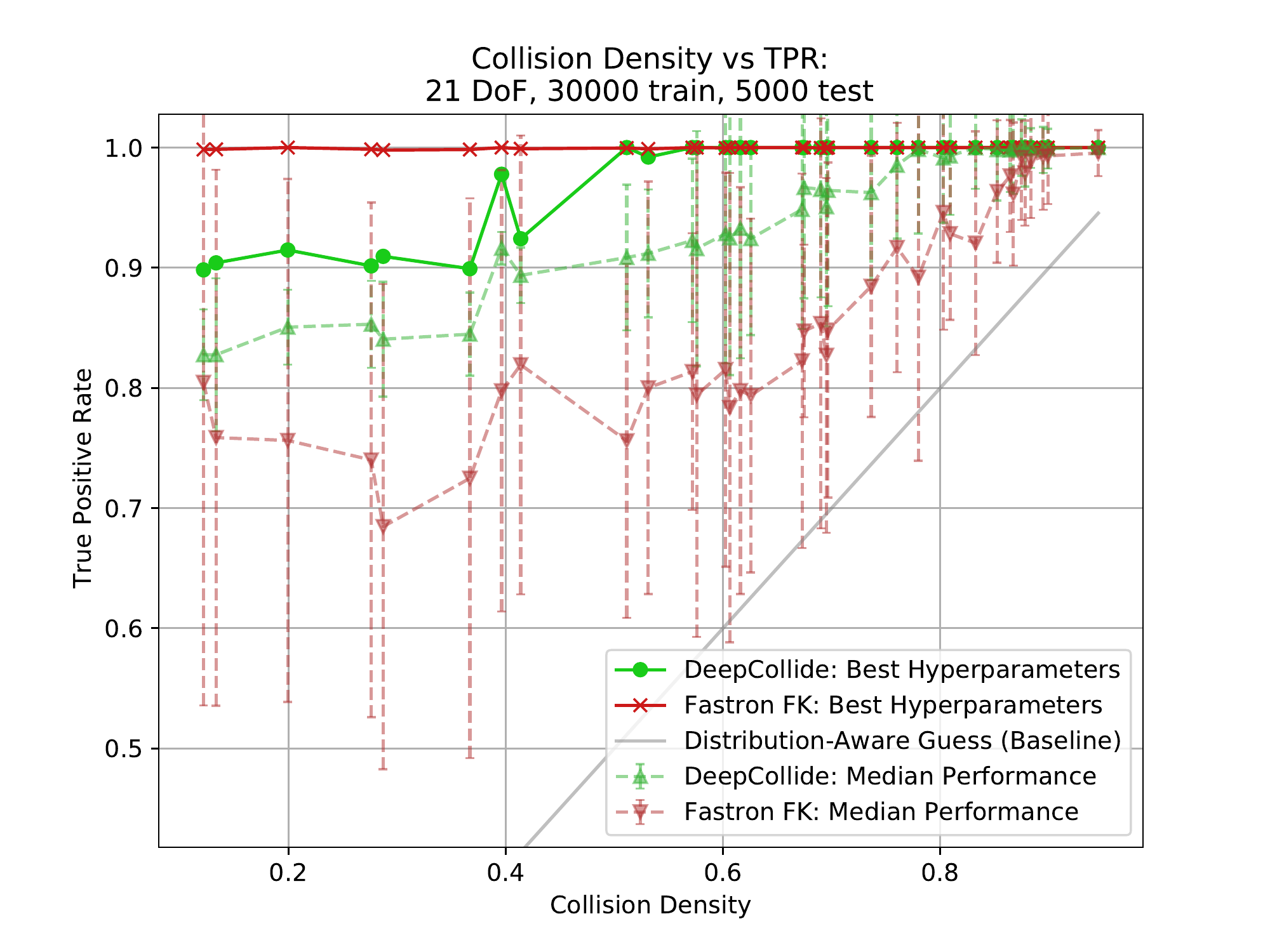}\label{fig:collision_density_tpr}}
    \hfill
    \subfloat[Collision Density vs. TNR]{\includegraphics[width=0.8\linewidth]{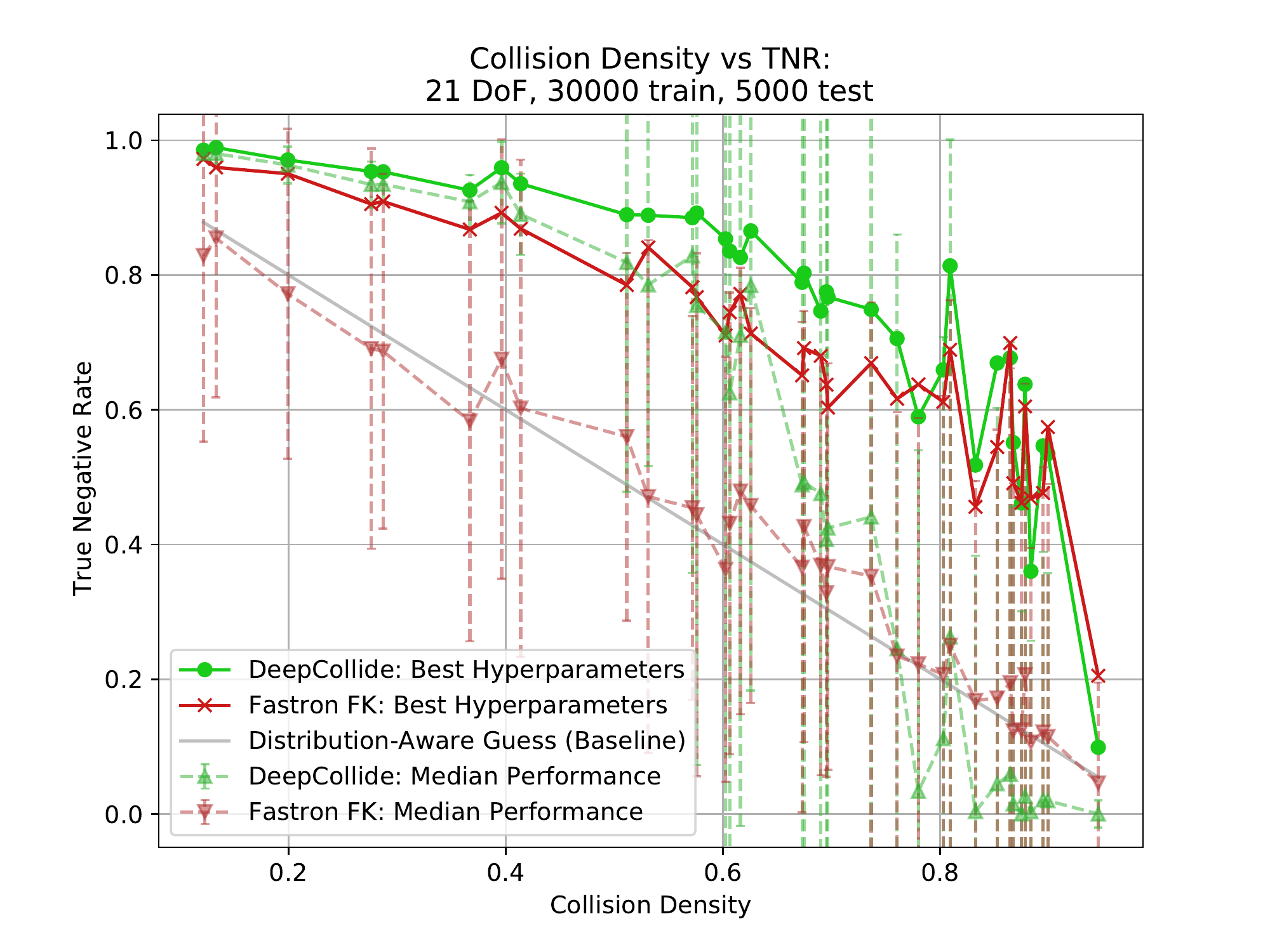}\label{fig:collision_density_tnr}}
    \hfill
    \subfloat[Collision Density vs. Accuracy]{\includegraphics[width=0.8\linewidth]{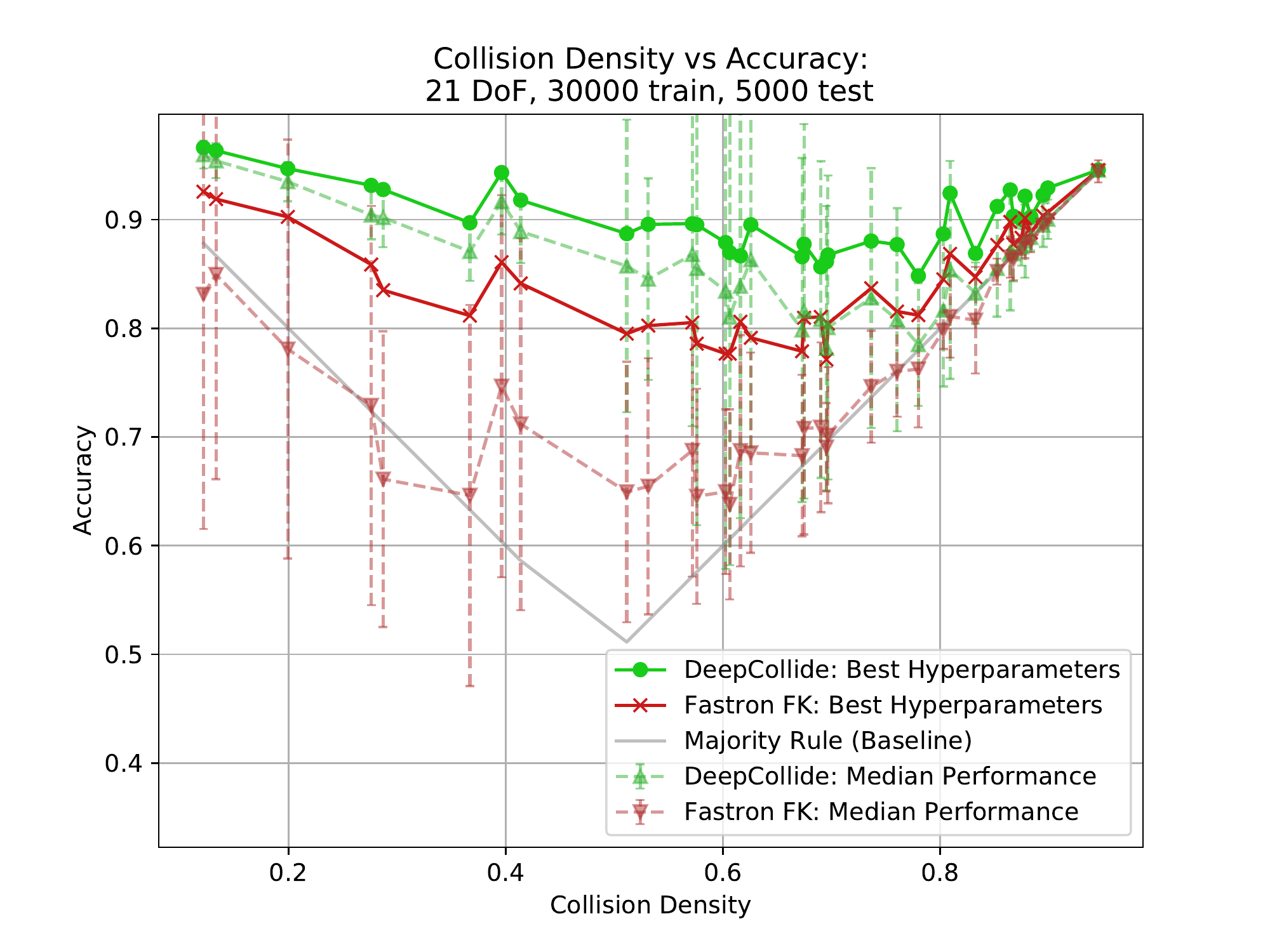}\label{fig:collision_density_accuracy}}
    \caption{\textbf{Impact of Collision Density on correctness.} 
    Interpretation is the same as Figure \ref{fig:dof_correctness_impact}, except that we are measuring collision density on the x-axis, and these graphs correspond to environments in Figure \ref{fig:collision_envs}, with only one environment per collision density.}
    \label{fig:collision_density_impact_correctness}
\end{figure}

\begin{figure}[h!]
    \centering
    \subfloat[Collision Density vs. Train Time]{\includegraphics[width=0.8\linewidth]{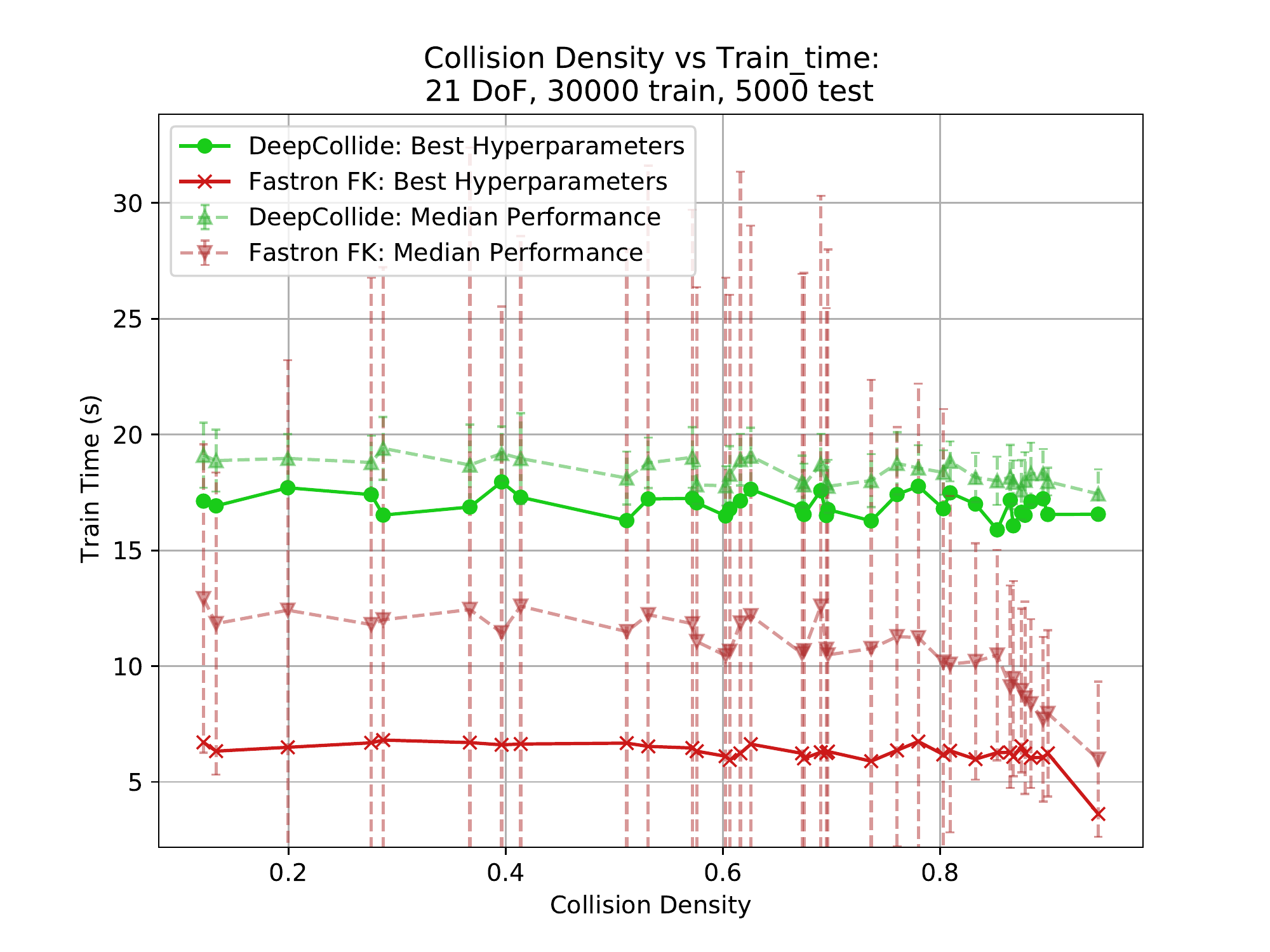}\label{fig:collision_density_traintime}}
    \hfill
    \subfloat[Collision Density vs. Test Time]{\includegraphics[width=0.8\linewidth]{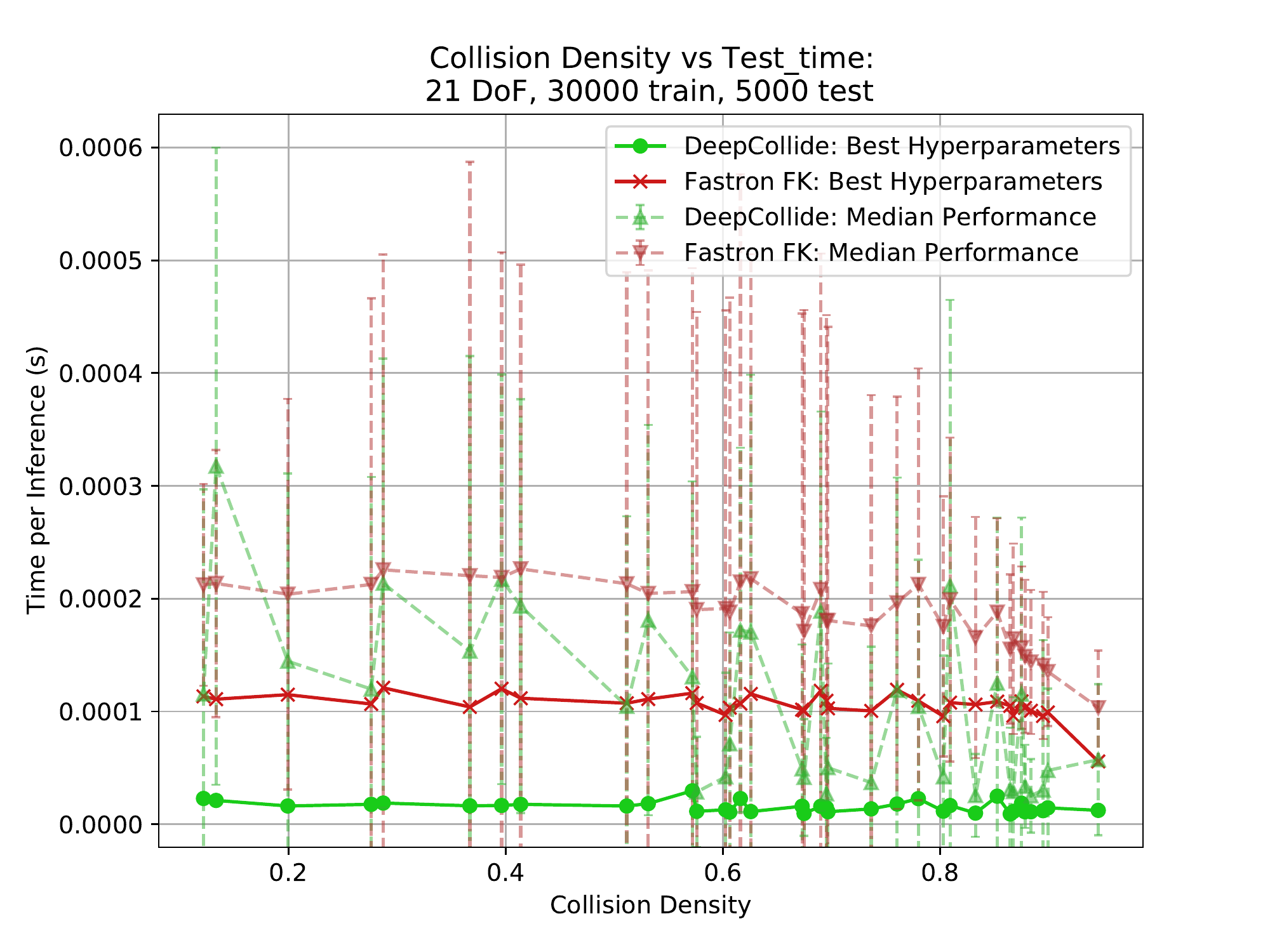}\label{fig:collision_density_testtime}}
    \caption{\textbf{Impact of Collision Density on speed.} Interpretation is the same as Figure \ref{fig:collision_density_impact_correctness}.
    }
    \label{fig:collision_density_impact_time}
\end{figure}

\section{Impact of Collision Density}

\subsection{Impact on Correctness}

Figure \ref{fig:collision_density_impact_correctness} shows the impact of collision density on model correctness. 
In general, we observe that as collision density increases, true positive rate increases for both models (Figure \ref{fig:collision_density_tpr}), and true negative rate decreases for both models (Figure \ref{fig:collision_density_tnr}). This is because we use uniform random sampling to gather our training dataset, so naturally, it is skewed with more collision samples as collision density increases, thereby biasing the model to predict collisions more frequently.

From Figure \ref{fig:collision_density_accuracy}, we see that DeepCollide consistently outperforms Fastron FK on accuracy. According to Figure \ref{fig:collision_density_tnr}, DeepCollide is also generally better than Fastron FK in regards to true negative rate. However, the results are less clear in regards to true positive rate, shown in Figure \ref{fig:collision_density_tpr}. We see that when favorable hyperparameters are chosen, Fastron FK performs better than DeepCollide, but in the median case (w.r.t. hyperparameter selection), DeepCollide is still better.

\subsection{Impact on Speed}

Figure \ref{fig:collision_density_impact_time} shows the impact of collision density on model speed. Firstly, we notice that overall, collision density does not significantly impact model speed in training or testing. The only exception is Fastron's slight speedup at the highest collision density shown. This may be due to Fastron's redundant support point removal mechanism, where it automatically removes points from the support set that would have a positive margin if they were excluded from the support set (more details in \cite{fastron_das2020learning}). In the high collision density case, fewer points may be needed to construct the support set, since most of the points have the collision label -- leading to quicker computations. 
Finally, in line with what we have seen so far, DeepCollide generally has quicker test time but slower train time than Fastron.

%% file: results/sample_size.tex
\begin{figure}[h!]
    \centering
    \subfloat[Sample Size vs. TPR (Baseline too low to be shown.)]{\includegraphics[width=0.8\linewidth]{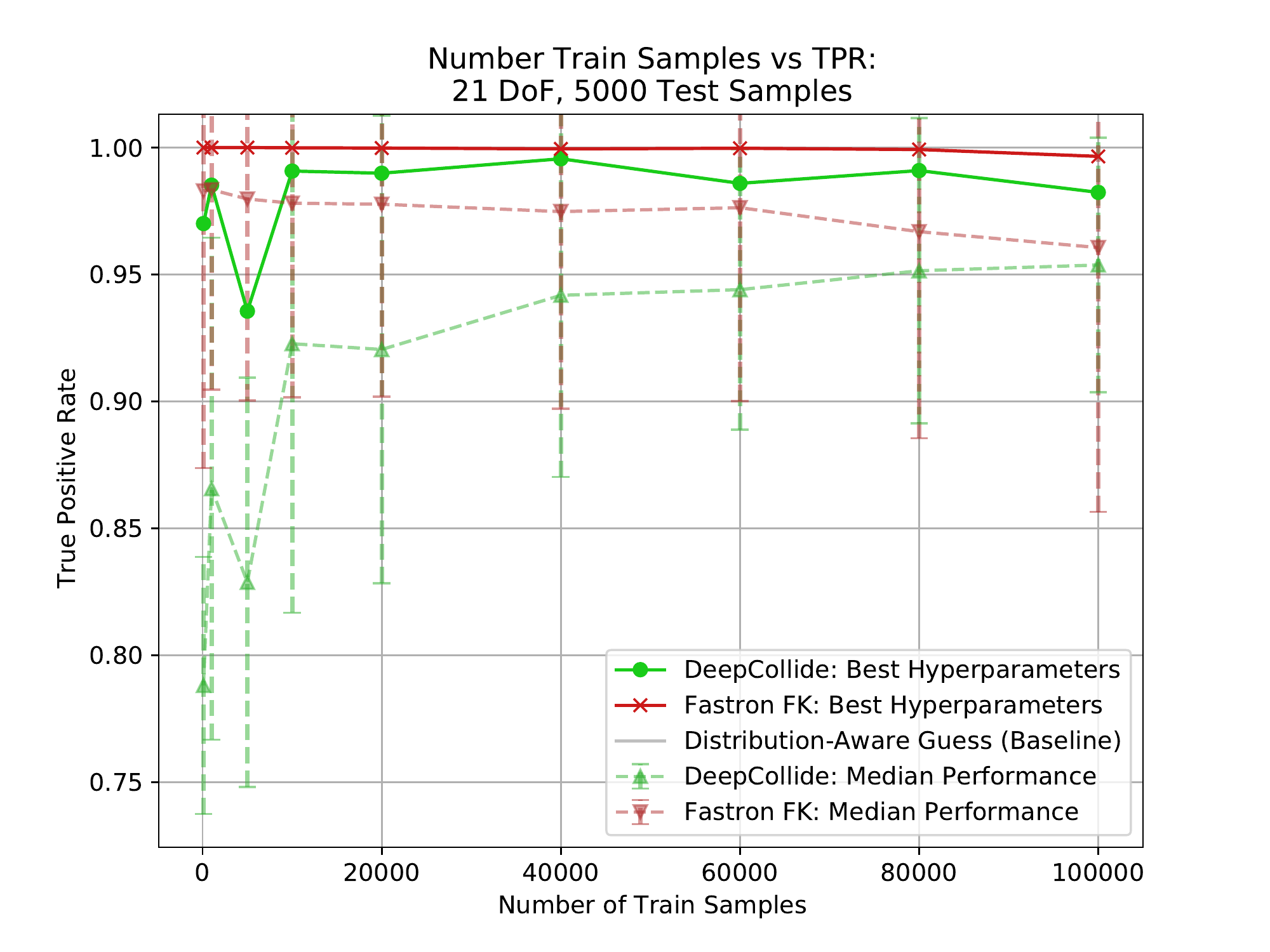}\label{fig:sample_size_tpr}}
    \hfill
    \subfloat[Sample Size vs. TNR]{\includegraphics[width=0.8\linewidth]{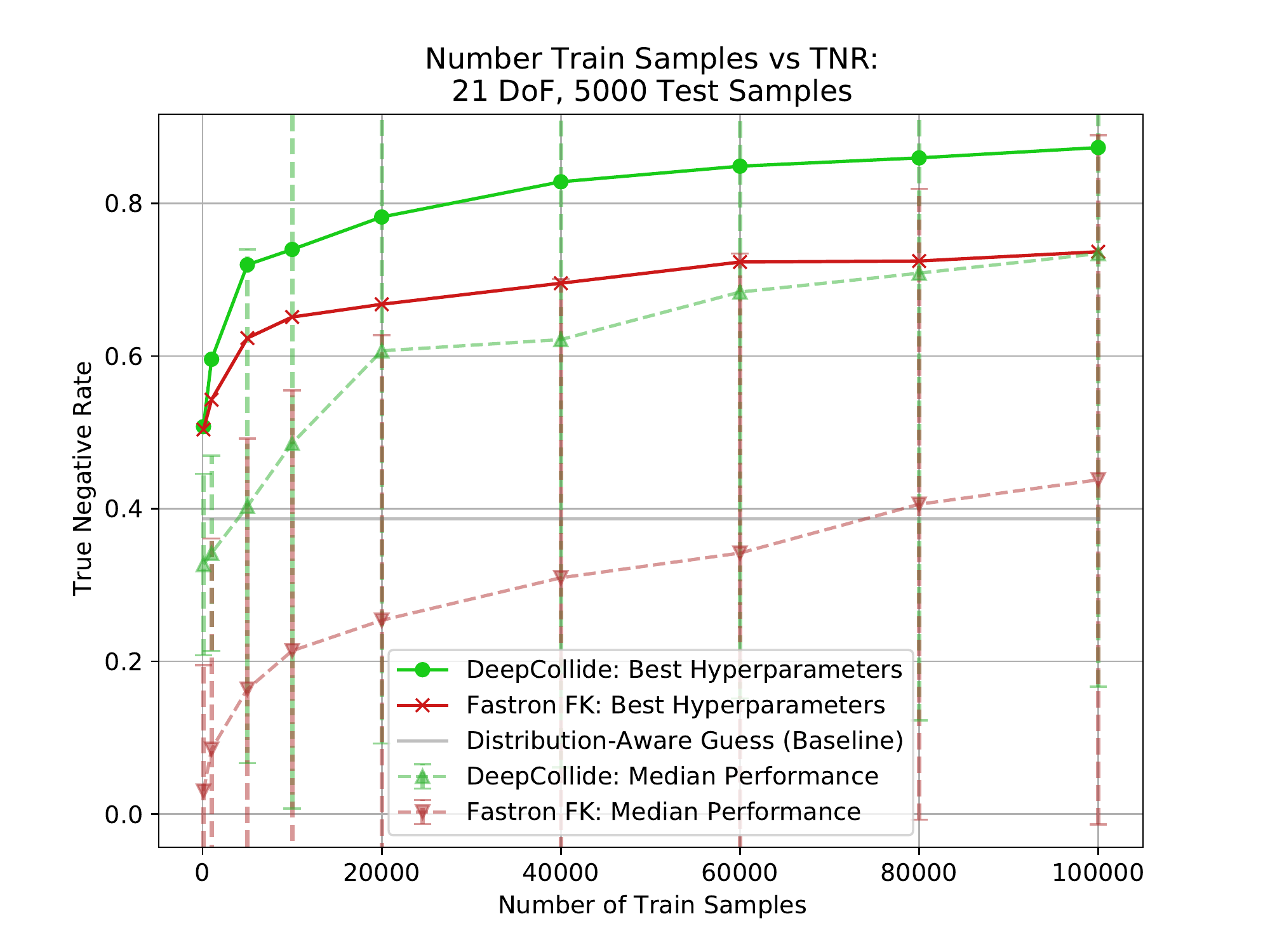}\label{fig:sample_size_tnr}}
    \hfill
    \subfloat[Sample Size vs. Accuracy]{\includegraphics[width=0.8\linewidth]{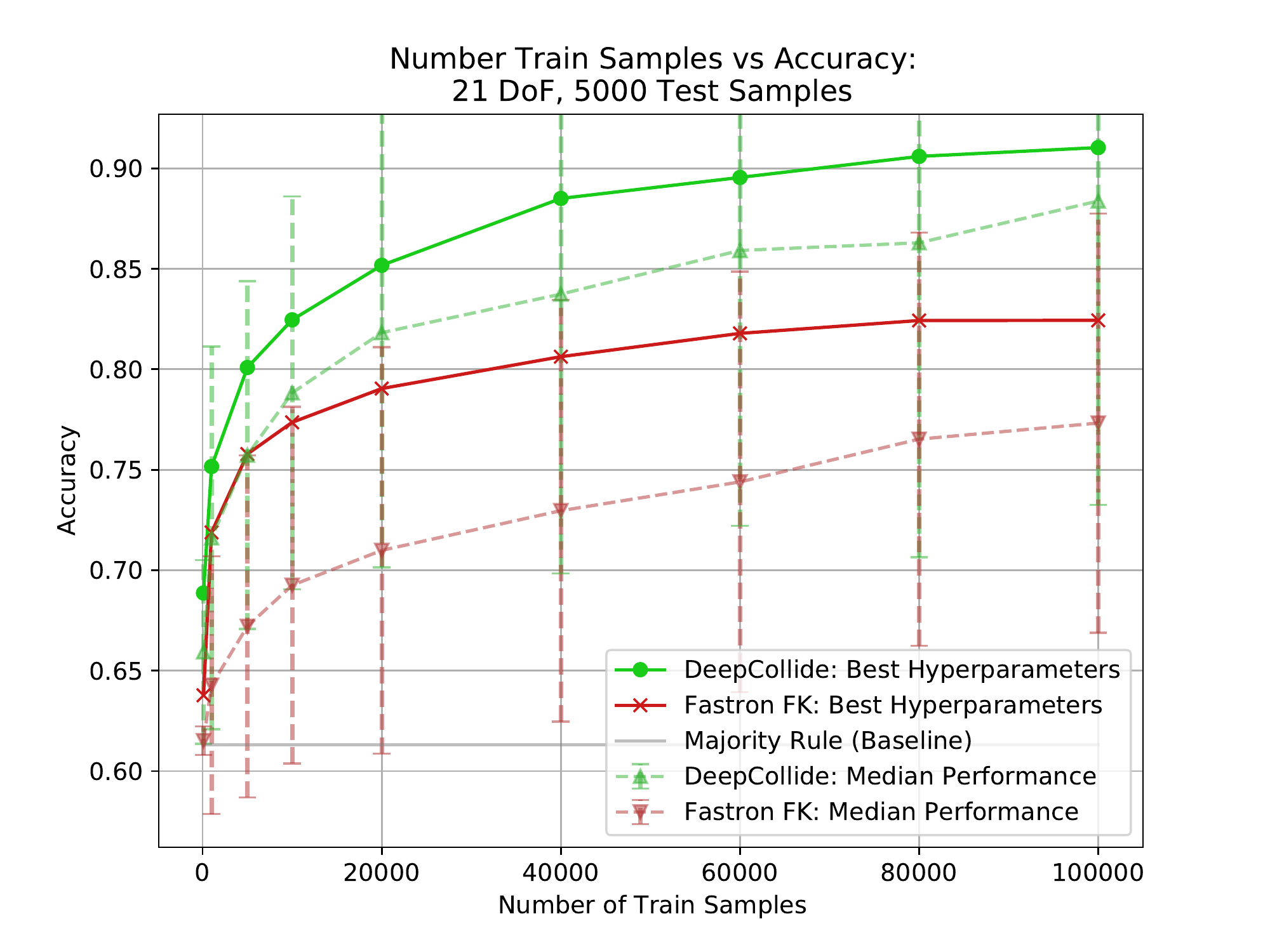}\label{fig:sample_size_accuracy}}
    \caption{\textbf{Impact of Sample Size on correctness.} The points on the graph show the average performance over three environments. Interpretation is the same as Figure \ref{fig:dof_correctness_impact}. 
    }
    \label{fig:sample_size_impact_correctness}
\end{figure}

\begin{figure}[h!]
    \centering
    \subfloat[Sample Size vs. Train Time]{\includegraphics[width=0.8\linewidth]{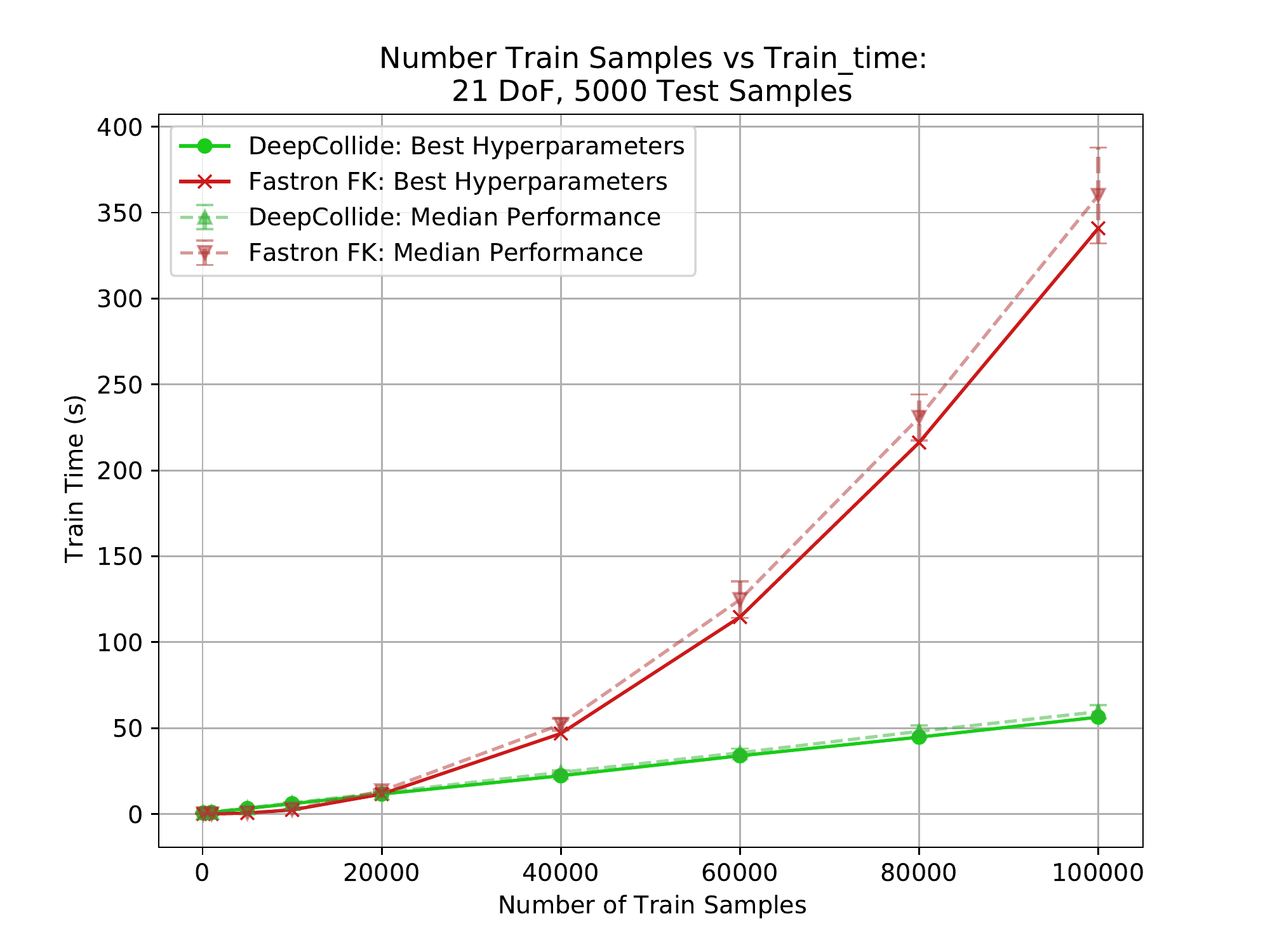}\label{fig:sample_size_traintime}}
    \hfill
    \subfloat[Sample Size vs. Test Time]{\includegraphics[width=0.8\linewidth]{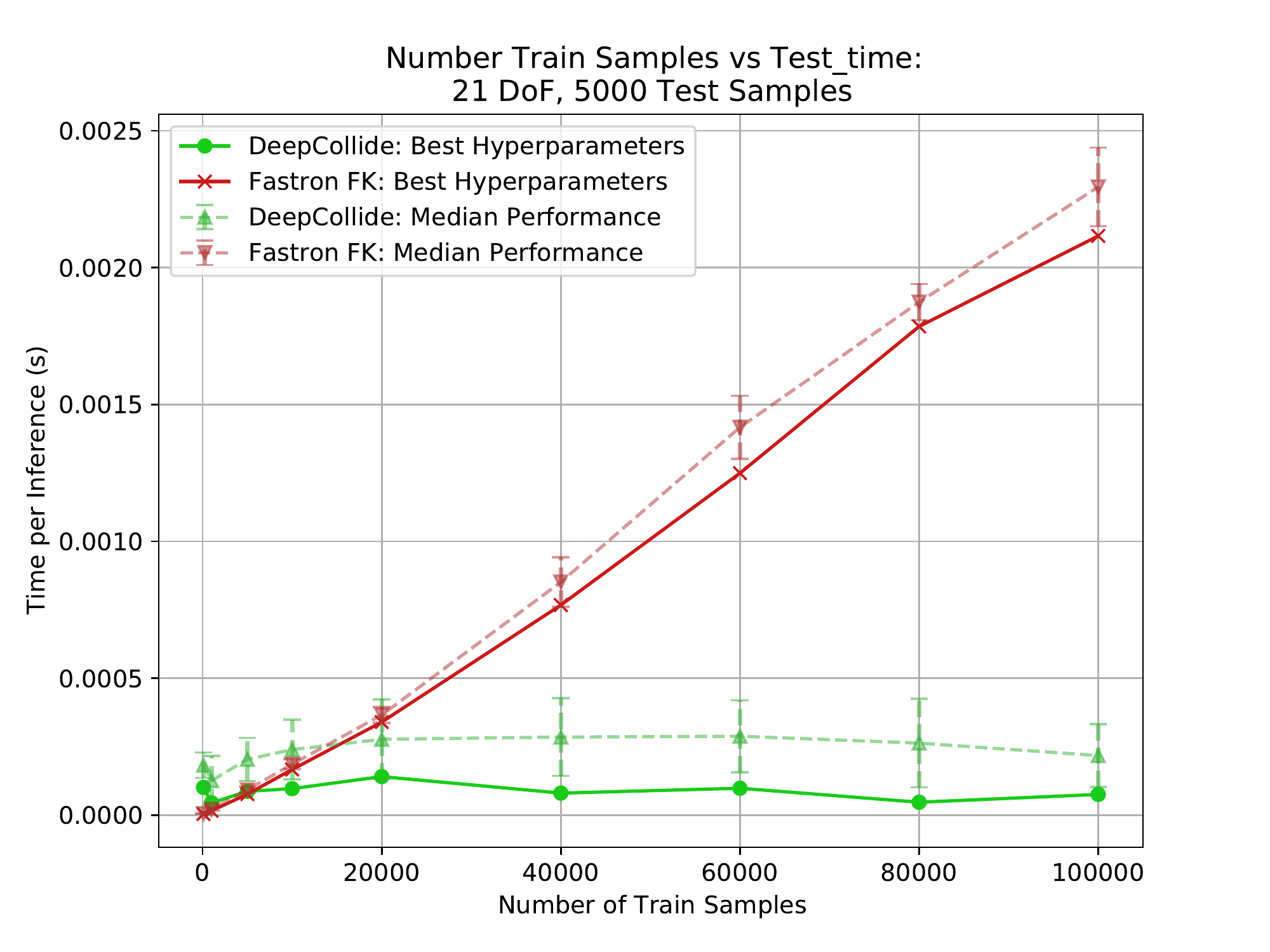}\label{fig:sample_size_testtime}}
    \caption{\textbf{Impact of Sample Size on time.} Interpretation is the same as Figure \ref{fig:sample_size_impact_correctness}. 
    }
    \label{fig:sample_size_impact_time}
\end{figure}

\section{Impact of Sample Size}\label{sec:training_data}

In measuring the impact of sample size, we set $\mathcal{I}_{max}$ to $50,000$ and $\mathcal{S}_{max}$ to $50,000$ for Fastron FK, so that it is equipped to process the large number of training points (up to $100,000$).

\subsection{Impact on Correctness}
Figure \ref{fig:sample_size_impact_correctness} shows the impact of train sample size on model correctness. Obviously, model performance gets better as we add more training points for both models. Zooming in, with regards to both accuracy (Figure \ref{fig:sample_size_accuracy}) and true negative rate (\ref{fig:sample_size_tnr}), DeepCollide clearly outperforms Fastron FK, regardless of sample size. With regards to true positive rate (Figure \ref{fig:sample_size_tpr}), Fastron FK outperforms DeepCollide over all sample sizes -- even so, both models have true positive rate above $90\%$ after we exceed $10,000$ samples, so the difference may not be so significant.

We also note that the support point and iteration limits ($\mathcal{S}_{max} = \mathcal{I}_{max} = 50,000$) on Fastron may theoretically diminish the accuracy benefits of more training data. However, as we see in the next subsection, Fastron already has a time scalability problem with respect to training data, and increasing those limits could only serve to exacerbate this.

\subsection{Impact on Speed}

Figure \ref{fig:sample_size_impact_time} shows the impact of train sample size on model speed. It is in line with Section \ref{sec:theoretical}'s theoretical expectations of time complexity. 
The implication of this is that DeepCollide scales much better than Fastron, with respect to training sample size. 

%% file: discussion_conclusion.tex
\section{Discussion, Limitations, and Conclusion}

In this work, we introduced DeepCollide, a scalable, lightweight neural network method for the rapid approximation of the collision detection function in high DoF. We showed that in a wide variety of high-DoF workspaces, it outperforms the state-of-the-art in regards to runtime-error tradeoff.
Furthermore, we found that DeepCollide scales well with regards to a variety of factors; including DoF, collision density, and training set size.

Our method could be most useful in situations where the obstacle and/or workspace geometry are not already known, but we do have sampled collision data from exploratory robot movements -- traditional geometric collision checkers will not work in these cases. It also has value in time-critical path-planning applications in complex but structured environments with repetitive tasks, such as multiple robotic arms collaborating inside a single factory cell. 

We also note that our method works in dynamically changing environments without having to make explicit updates to the model. In the experiments where we have multiple robotic arms, we can consider each arm to be a dynamically moving obstacle to the other arms -- this shows that as long as the moving obstacle is accounted for in a degree of freedom (\textit{i.e.}, as an input to the network), our method works. Other methods \cite{fastron_das2020learning} were designed to work in dynamically changing environments, but they require explicit updates to the model.

We focus our work on collision detection, which is part of the larger motion planning pipeline \cite{kuffner2000rrt, prm_kavraki1996probabilistic, fastron_das2020learning}. We leave explicit motion planning experiments to future work.


Another key limitation of this study is that the data collected to train the models was uniformly sampled. Future work will consider adaptive sampling approaches where samples are collected predominantly near obstacle boundaries or in areas of interest, as well as active-learning approaches where samples are collected to deliberately mitigate uncertainty in the learned models.

Finally, if GPUs are available during inference time, we can further increase the speed of our method.